\documentclass[lettersize,journal]{IEEEtran}
\usepackage{amsmath,amsfonts}
\usepackage{algorithmic}
\usepackage{algorithm}
\usepackage{array}
\usepackage{textcomp}
\usepackage{stfloats}
\usepackage{url}
\usepackage{verbatim}
\usepackage{graphicx}
\usepackage{multirow}
\usepackage{subfigure}
\usepackage[nocompress]{cite}
\usepackage{color}
\usepackage{xcolor}

\usepackage{hyperref}
\hypersetup{
colorlinks=true,
linkcolor=blue,
filecolor=blue,
urlcolor=blue,
citecolor=blue,
}
\usepackage{CJKutf8}

\begin{document}

\title{Speak the Same Language: Global LiDAR Registration on BIM Using Pose Hough Transform}

% \title{Speak the Same Language: LiDAR to BIM Alignment Using Pose Hough Transform}

% \author{Zhijian Qiao$^\dag$, Haoming Huang$^\dag$, Chuhao Liu, Shaojie Shen, Jack C.P. Cheng, Fumin Zhang and Huan Yin
\author{Zhijian Qiao$^\dag$, Haoming Huang$^\dag$, Chuhao Liu, Zehuan Yu, Shaojie Shen, Fumin Zhang and Huan Yin
        % <-this % stops a space
\thanks{$^\dag$ Zhijian Qiao and Haoming Huang contributed equally to this work. }% <-this % stops a space
\thanks{Zhijian Qiao, Chuhao Liu, Zehuan Yu, Shaojie Shen, Fumin Zhang, and Huan Yin are with the Department of Electronic and Computer Engineering, Hong Kong University of Science and Technology, Hong Kong SAR. }
\thanks{Haoming Huang is with the Division of Emerging Interdisciplinary Areas, Hong Kong University of Science and Technology, Hong Kong SAR. }
% \thanks{Jack Chin Pang Cheng is with the Department of Civil and Environmental Engineering, Hong Kong University of Science and Technology, Hong Kong SAR. }
\thanks{This work was supported in part by the HKUST-DJI Joint Innovation Laboratory, in part by the HKUST-BDR Joint Research Institute (Grant No. OKT25EG04), and in part by the Hong Kong Center for Construction Robotics (InnoHK center supported by Hong Kong ITC).}
\thanks{Corresponding author: Huan Yin}
}

% The paper headers
% \markboth{Journal of \LaTeX\ Class Files,~Vol.~14, No.~8, August~2021}%
\markboth{Accepted by the Special Issue on Automation and AI in Construction and Building, IEEE T-ASE}%
{Shell \MakeLowercase{\textit{et al.}}: A Sample Article Using IEEEtran.cls for IEEE Journals}

% \IEEEpubid{0000--0000/00\$00.00~\copyright~2021 IEEE}
% Remember, if you use this you must call \IEEEpubidadjcol in the second
% column for its text to clear the IEEEpubid mark.

\maketitle

\begin{abstract}
Light detection and ranging (LiDAR) point clouds and building information modeling (BIM) represent two distinct data modalities in the fields of robot perception and construction. These modalities originate from different sources and are associated with unique reference frames. The primary goal of this study is to align these modalities within a shared reference frame using a global registration approach, effectively enabling them to ``speak the same language''. To achieve this, we propose a cross-modality registration method, spanning from the front end to the back end. At the front end, we extract triangle descriptors by identifying walls and intersected corners, enabling the matching of corner triplets with a complexity independent of the BIM's size. For the back-end transformation estimation, we utilize the Hough transform to map the matched triplets to the transformation space and introduce a hierarchical voting mechanism to hypothesize multiple pose candidates. The final transformation is then verified using our designed occupancy-aware scoring method. To assess the effectiveness of our approach, we conducted real-world multi-session experiments in a large-scale university building, employing two different types of LiDAR sensors. We make the collected datasets and codes publicly available to benefit the community.
\end{abstract}

\def\abstractname{Note to Practitioners}
\begin{abstract}
Our proposed registration method leverages walls and corners as shared features between LiDAR and BIM data, making it particularly well-suited for scenarios with well-defined structural layouts. Accumulating a larger LiDAR submap provides richer structural information, which further aids in achieving accurate alignment. To optimize computational efficiency, we recommend constructing the descriptor database offline and loading it during runtime, enabling a theoretical retrieval complexity of $O(1)$. Despite its advantages, our approach has certain limitations. First, it primarily focuses on planar structures, which limits its effectiveness in utilizing non-planar features. Second, the method may underperform in cases where significant deviations exist between the as-designed BIM and as-is LiDAR data. Lastly, in ambiguous scenarios, such as long corridors or similar layouts within the same or across different floors, our method may struggle to verify the correct transformation among candidates. To address these challenges, incorporating additional information, particularly semantic cues such as floor numbers, room numbers, and room types, could enhance its robustness and reliability .
\end{abstract}

\begin{IEEEkeywords}
Point cloud registration, LiDAR, Building information modeling, Hough transform
\end{IEEEkeywords}

\section*{Supplementary Materials}

{Video} is available \href{https://www.youtube.com/watch?v=SWbnsaRyL-M}{online}.

Code will be released at \href{https://github.com/HKUST-Aerial-Robotics/LiDAR2BIM-Registration.git}{https://github.com/HKUST-Aerial-Robotics/LiDAR2BIM-Registration.git}.

The LiBIM-UST dataset of this study has been upgraded to the SLABIM dataset, available at \href{https://github.com/HKUST-Aerial-Robotics/SLABIM}{https://github.com/HKUST-Aerial-Robotics/SLABIM}.

\section{Introduction}

% \href{https://www.ieee-ras.org/images/publications/t-ase/2024/T-ASE_SI-Proposal_Automation_in_Construction_CFP_final.pdf}{SPECIAL ISSUE LINK}

\IEEEPARstart{T}{hree}-dimensional (3D) point cloud data provides accurate modeling of the indoor environment. Its application extends across various phases in the construction, encompassing typical uses such as geometry quality inspection and real-time visualization~\cite{wang2019applications}. Traditionally, obtaining precise point clouds often relied on stationary laser scanners. In the last ten years, mobile LiDAR mapping has emerged as a well-established technique within the robotics community~\cite{xu2022fast}, like centimeter-level simultaneous localization and mapping (SLAM). This approach facilitates the rapid acquisition of 3D point clouds using mobile robotic platforms, thus advancing 3D point cloud-based applications in both robotics and construction.

In the architectural, engineering, and construction (AEC) industry, the utilization of 3D point clouds frequently involves the as-designed building data, e.g., visualizing elements of BIM with augmented reality (AR) techniques or comparing LiDAR to BIM for quality management. Notably, a rigid transformation inherently exists between the as-built source frame and the as-designed reference frame, making \textit{frame alignment} essential for integrating distinct data sources into a unified spatial space for subsequent analysis. In other words, downstream comparisons and analyses involving BIM cannot be performed without proper frame alignment. In commercial software, users manually set the transformation to align the data, and manual tuning is necessary to ensure alignment quality. Regarding long-term robotic operations, external infrastructures are typically required to calibrate the different frames, like employing tags and markers in the building. However, these processes entail much human labor and additional costs, restricting the adoption of LiDAR point clouds in the construction industry, especially in large-scale environments. This frame alignment challenge also arises for mobile robots during global localization, where the robot must align itself with a pre-built map, such as BIM, as an initial step~\cite{yin2024survey}.

In this study, we propose an automatic alignment approach for LiDAR point clouds and BIM using a \textit{global point cloud registration} approach, without the requirements of external infrastructures or manual verification. Global point cloud registration has been studied in the field of robotics and computer vision, such as LiDAR-only registration~\cite{lim2024quatro++,qiao2024g3reg}, and indoor RGBD alignment~\cite{halber2017fine}. Compared to these tasks, we face three main challenges: \textit{1) the inherent difference in modality between LiDAR point clouds and BIM, which introduces significant difficulties in front-end feature extraction and data association; 2) repetitive patterns in indoor environments, which can result in ambiguity for pose estimation; 3) the deviation between as-built and as-designed data, making reliable data association challenging to achieve.} To address these challenges, our contributions encompass both front-end and back-end processes. We employ walls as the fundamental representations of BIM and LiDAR data, enabling the extraction of corner points and point-level correspondences (data association). Multiple pose candidates are subsequently obtained through voting based on the Hough transform and a hierarchical voting scheme. These pose candidates are verified using our proposed occupancy-aware score, leading to the final pose estimation. We validate our approach through experiments conducted on real-world data collected from a large-scale university building. The dataset includes multi-session LiDAR point clouds collected from mobile mapping platforms with two different LiDAR sensor types. Beyond its application in construction, the proposed global registration method can also be incorporated into robot navigation systems that leverage BIM as a navigational map~\cite{hendrikx2021connecting,yin2023semantic}.

Overall, this study focuses on globally registering LiDAR data to BIM, and we summarize the contributions as follows:

\begin{itemize}
    \item We design and develop a complete registration framework capable of automatically and globally aligning LiDAR point clouds with BIM in one shot.
    \item A comprehensive descriptor with $O(1)$ retrieval complexity is proposed for cross-modality data association, effectively encoding basic walls and corners in structural environments.
    \item A Hough transform scheme with hierarchical voting is introduced to hypothesize multiple pose candidates, followed by verifying the optimal transformation using an occupancy-aware score, improving both robustness and accuracy.
    \item We conduct real-world experiments using two different LiDAR sensors in a large-scale university building. Additionally, we plan to open-source the LiBIM-UST dataset and codes to benefit the research community.
\end{itemize}

The structure of this paper is organized as follows: Section~\ref{sec:related} presents the related work on BIM-aided robotic mapping and global point cloud registration; Section~\ref{sec:overview} introduces the system overview. Section~\ref{sec:frontend} details the the feature extraction and correspondence generation, following the pose Hough transform and voting in Section~\ref{sec:backend}. We validate the proposed method in the real world in Section~\ref{sec:experiments}. A conclusion is summarized in Section~\ref{sec:conclusion}.

\section{Related Work}
\label{sec:related}

In this section, we first provide related works on robotic sensing and BIM. We then dive into the specific topic of point cloud registration, emphasizing the importance of point descriptors in this context.

\subsection{Robots Meet BIM}

BIM serves as a digital representation of buildings and other physical assets in the AEC industry. BIM generally encompasses comprehensive 3D spatial information, making it suitable for aligning robotic sensing data with the BIM model. This alignment process can effectively reveal deviations between the as-built and as-designed states, thereby benefiting the AEC industry. Conversely, BIM could also serve as a spatial map that facilitates robot localization and planning, eliminating the need for traditional mapping approaches. By leveraging the information provided by BIM, robots can navigate and operate within the environment without relying on explicit mapping procedures. This mapping-free characteristic of BIM enhances robot navigation capabilities. The subsequent paragraphs will delve into the relevant works conducted in these two aspects, namely aligning robot sensings to BIM for deviation analysis and utilizing BIM as a spatial map for robot navigation.

Visual data is highly informative in nature. Han \textit{et al.} \cite{asadi2019real} introduce a technique where vanishing lines and points are detected in images and subsequently projected onto the BIM for the purpose of progress monitoring. Similarly, Chen \textit{et al.} \cite{chen2022align} propose a method that enables the alignment of photometric point clouds with BIM, thereby achieving precise camera localization, commonly referred to as the "align-to-locate" approach. It is worth noting that a single 2D image cannot inherently provide depth information, which restricts the deployment on image-to-BIM. While stereo vision can offer depth perception~\cite{hartley2003multiple}, its suitability for large-scale alignment is limited. In light of these challenges, our study proposes the utilization of LiDAR point clouds as the input, facilitating direct 3D modeling of the environment.

In the robotics community, an existing recognition is that for long-term robots operating in stable environments, the mapping process of SLAM can be redundant as the generated map remains largely unchanged over time. As an alternative, utilizing BIM or low-level floorplans has emerged as a potential solution for long-term robot navigation. One notable approach is FloorPlanNet, proposed by Feng \textit{et al.} \cite{feng2023floorplannet}, which segments rooms from point cloud data and aligns them with a floor plan. The core architecture of FloorPlanNet is based on a graph neural network for effective feature learning. Similarly, Zimmerman \textit{et al.} \cite{zimmerman2023constructing} introduce a semantic visibility model integrated into a particle filter, enhancing long-term robot indoor localization. These works leverage floorplans as 2D maps, while BIM is a good choice for providing 3D point cloud maps. Yin \textit{et al.} \cite{yin2023semantic} generated a semantic-metric point cloud map from a BIM and designed a semantic-aided iterative closest point (ICP) algorithm for robot localization. This method effectively localizes a robot equipped with a moving sensor within the BIM environment. While the aforementioned works focus on point-level operations, there is also a promising direction in building instance-level robot navigation. In the study by Shaheer \textit{et al.} \cite{shaheer2023robot}, a novel map called S-Graph is designed based on BIM. The S-Graph consists of three-layered hierarchical representations, and a particle filter-based approach is employed to support robot navigation at the instance level. 

All the works above aim to align observed data to a floorplan-based or BIM-based map, in which point cloud registration is an indispensable function. This will be discussed in the following subsection.

% Huan Yin~\cite{yin2022towards,yin2023semantic}

% BIM can be integrated with a Lidar-SLAM system in various angles. 

% The room features are projected on image plane to encode and spatial information is not effected used in the graph embedding process.

% Compare with previous Lidar registration methods in a BIM, we are more focused on real-world 3D registration performance. All of our experiments are based on real-world data, and involve large-scale buildings in multiple Lidar devices. \lchcomment{may be inaccurate. need to confirm our advantages.}

% Overall, leveraging BIM into the eob

% \yhcomment{why construction need alignment to BIM} \lchcomment{answered in the first paragraph} 

\subsection{3D Point Cloud Registration and Advances}

% Local point cloud registration ICP~\cite{besl1992method} and NDT~\cite{magnusson2007scan} iteratively find the nearest neighbor correspondences and optimize the relative pose. Although they are precise in registering local point clouds, they cannot guarantee the registration in large pose differences. 
% % i.e., for global registration

% Global Registration Survey~\cite{yin2024survey} Outram by Pengyu Yin~\cite{yin2023outram}, Teaser~\cite{yang2020teaser} G3Reg~\cite{qiao2023g3reg}

% Deep Learning-based methods and challenges, and why we do not take deep learning in this task

% Recent advances for registration, Cross-Modality Registration CAS~\cite{zhao2023accurate} Semantic-aided~\cite{liu2019global} \cite{yin2023outram}

Point cloud registration, like ICP~\cite{besl1992method} and normal distributions transform (NDT)~\cite{magnusson2007scan}, focuses on local point-level alignment with a good initial guess. Global point cloud registration aims at solving pose estimation from scratch~\cite{yin2024survey}. The global point cloud registration can be categorized into two primary models: correspondence-free\cite{bernreiter2021phaser} and correspondence-based. This study is in a correspondence-based manner. We mainly focus on point descriptors to provide correspondences and recent advances in this task.

% \subsubsection{Feature Matching}
Traditional 3D descriptors like Fast Point Feature Histogram (FPFH)~\cite{rusu2009fpfh} compute rotation-invariant features such as distances and angles between points and normals. Then, it searches correspondences on a k-d tree constructed from the FPFH features. Although it performs well in environments with rich geometric features, it degenerates in scenes with ambiguous structures. With the development of 3D representation learning such as PointNet~\cite{qi2017pointnet} and KPConv~\cite{thomas2019kpconv}, various of learning-based methods are proposed to encode and match point cloud features. Deep Closet Point (DCP)~\cite{wang2019deep} encodes 3D point features and incorporates self-attention and cross-attention layers on the encoded features. Point correspondences can be searched by computing the all-to-all feature similarities. 
After that, GeoTransformer~\cite{qin2022geotransformer} utilizes KPConv to encode hierarchical 3D point features and use optimal transport for correspondence estimation. Although DCP and GeoTransformer have achieved promising results on small-scale benchmarks~\cite{wu2015shapenet,zeng20173dmatch}, they face two main challenges for LiDAR global registration with BIM. Firstly, these environments contain numerous similar 3D structures, especially surface planes, making it difficult to distinguish and match corresponding elements accurately. Secondly, the differences in modalities between BIM and LiDAR data create a large domain gap, leading to ambiguous encoding of the 3D point cloud by the 3D backbone. Additionally, learning-based methods incorporating multiple attention layers consume a large amount of memory, making them impractical for use in large-scale environments with current computing platforms.

\begin{figure*}[t]
  \centering
  \includegraphics[width=0.95\textwidth]{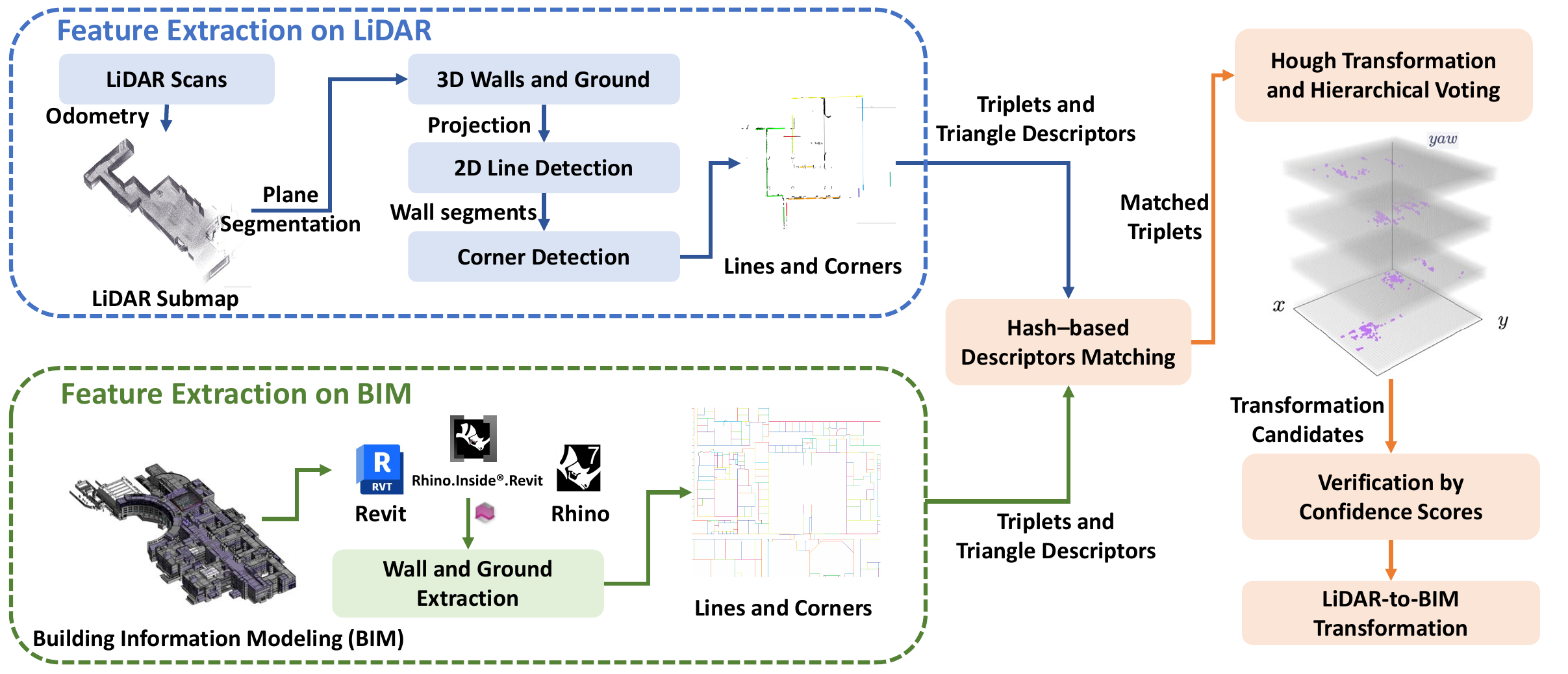}
  \caption{The pipeline of our proposed method begins with the accumulation of LiDAR scans to build submaps using LiDAR odometry (Section \ref{sec:submap}). From these submaps, wall, corner, and ground features are extracted (Section \ref{sec:feature}) to construct triplets and generate triangle descriptors (Section \ref{sec:descriptor}). For BIM preprocessing, specialized software tools are employed to extract wall and corner features (Section \ref{sec:bim_de}). The triangle descriptors are efficiently matched using a hash algorithm, enabling one-shot registration across the entire BIM. Based on the matched triplets, transformation candidates are generated, and a pose Hough transform combined with a hierarchical voting procedure is applied to identify the most plausible transformation candidates (Section \ref{sec:hough}). Finally, an occupancy-aware confidence score is introduced to determine the optimal transformation (Section \ref{sec:verify}).}
  \label{fig:pipeline}
\end{figure*}

% : coarse-level and fine-level point features. It incorporates attention layers on coarse-level features, enriching the feature descriptive in a larger perception field. Then, it computes the similarity matrix at each level of features. The matched feature can be found using an optimal transport algorithm like Sinkhorn~\cite{peyre2019sinkhorn}. Thanks to the hierarchical feature embeddings and coarse-to-fine match strategy, GeoTransformer guarantees a higher inline ratio and significantly specifies the nodes in attention layers and point-matching step. 

% However, a room is matched by searching the nearest room and preventing it from being used in global registration.

The 3D point descriptor can also be constructed along with semantic information. Given an RGB-D sequence as input, Kimera by Rosinol \textit{et al.}~\cite{rosinol2021kimera} constructs a hierarchical scene graph in a large-scale indoor environment. Based on Kimera, Hydra~\cite{hughes2022hydra} encodes the semantic objects and rooms as hierarchical features. Some related methods reconstruct a semantic graph and encode the graph topology using random walk descriptors. Outram by Yin \textit{et al.}~\cite{yin2024outram} proposes a triangulated 3D scene graph to search correspondences. BoxGraph by Pramatarov \textit{et al.}~\cite{pramatarov2022boxgraph} clusters the semantic point cloud into instances. It matches each pair of instances while considering their shape similarity and semantic label. In~\cite{zimmerman2023constructing}, the floorplan-based method extracts the semantic objects in the 2D floorplan and incorporates the objects into the Monte Carlo Localization model. These semantic-aided approaches are hard to apply in our task registration because the semantic information significantly differs between a BIM model and a LiDAR data, especially when a large deviation exists between the as-designed and as-built in the semantic level. BIM mainly involves structural elements such as walls and rooms, while the methods above detect furniture and small-size objects in LiDAR or RGB-D maps. The inconsistent semantic descriptors pose serious challenges in matching them across modalities.

% Random sample consensus (RANSAC)~\cite{fischler1981random} is an iterative method that have been used in image matching and point cloud registration. In recent years, another graph-theoretic approach is solving maximum clique~\cite{yang2020teaser,lusk2021clipper,zhang2023mac}. It requests that the correspondence points be mutually consistent or treated as outliers and pruned. The maximum clique method demonstrates promising results in various environments. TEASER~\cite{yang2020teaser} has rejected most of the outliers during the registration. On the other hand, the maximum clique problem is known as NP-hard. It is time-consuming once the number of nodes is on a large scale. With a set of point correspondences, the pose estimation can be formulated as a least-square problem. The optimization can be solved in a closed-form \cite{horn1987closed,censi2008icp}with zero or low outlier ratio. To estimate the pose with measured outliers, a robust pose estimator is required \cite{yang2020teaser,yang2020gnc}. 

% Teaser \cite{yang2020teaser} has achieved robust pose estimation in $70-80\%$ outlier ratio.

% Moreover, the scene graph reconstructed by Kimera can be noise and hard to be generalized to real-world scenario. 

% \begin{figure*}[t]
%   \centering
%   \includegraphics[width=\textwidth]{figs/overview.png}
%   \caption{Example of a wide figure spanning across two columns. \yhcomment{NEED RE-DESIGN. Like Dr. Jiarong Lin's Style.}}
%   \label{fig:overview}
% \end{figure*}

\section{System Overview}
\label{sec:overview}

The primary objective of this paper is to achieve robust and global registration between BIM and LiDAR scans. The core challenge arises from the inherent modality difference, which manifests in variations in structure, density, and information content. To tackle this, we have developed a pipeline that first constructs correspondences by identifying common features in the front end and then performs pose (transformation) estimation based on these correspondences in the back end.

The architecture of our proposed pipeline is illustrated in Figure~\ref{fig:pipeline}. The process begins with data preprocessing, as outlined in Section~\ref{sec:preprocess}, where consecutive LiDAR scans are accumulated into LiDAR submaps using the relative pose estimates from the LiDAR Odometry. The subsequent extraction of common semantic features, specifically walls and corners, from the submaps and BIM is discussed in Section~\ref{sec:feature}. These features are then associated through the construction and matching of a novel triangle descriptor, as detailed in Section~\ref{sec:descriptor}.

A transformation estimation technique using the Hough transform and voting is introduced in Section~\ref{sec:hough}, which utilizes corner-wise correspondences formed by the triangle descriptor based matching. This estimator generates multiple candidate transformations through a voting mechanism. The multi-candidate approach is designed to address the challenge of potential true transformations being mistakenly discarded due to repetitive corner patterns within the building structures. The correct transformation is subsequently selected using a verification scheme, which identifies the candidate that most accurately aligns the submap with the BIM, as detailed in Section~\ref{sec:verify}.

\section{Front-end: Walls, Corners and Descriptors}
\label{sec:frontend}

In this section, we establish correspondences between BIM and LiDAR submaps by extracting common features: walls and corners. The proposed pipeline consists of three main steps, as illustrated in Figure~\ref{fig:pipeline}. First, sequential LiDAR point clouds are accumulated into a submap $\mathcal{S}$ to improve the field of view, and the BIM is processed for element extraction $\{\mathcal{B}\}$ (Section~\ref{sec:preprocess}). Second, walls and corners are detected and extracted in both $\mathcal{S}$ and $\{\mathcal{B}\}$ (Section~\ref{sec:feature}). Finally, corners are used to construct triangle descriptor dictionaries for point-level data associations (Section~\ref{sec:descriptor}).

\subsection{LiDAR Submaps and BIM}
\label{sec:preprocess}

\subsubsection{Building LiDAR Submaps} 
\label{sec:submap}

The proposed approach for global LiDAR registration on BIM utilizes LiDAR submaps (also referred to as local maps). The preference for submaps over individual scans is based on two key reasons. First, submaps provide a more uniform point density than a single LiDAR scan, enhancing the accuracy of feature extraction for elements, as detailed in Section~\ref{sec:feature}. Second, submaps cover a larger area compared to single scans, reducing ambiguity caused by limited viewpoint—particularly in indoor environments where views are frequently occluded—and thereby increasing the robustness of the registration process.

% The density and size of a single LiDAR scan and a BIM model often differ significantly due to modality differences. For effective construction monitoring, denser LiDAR data are preferable to capture detailed geometric information. Conversely, to reduce the complexity of the problem, preprocessing of LiDAR scans and BIM is necessary to align them in size and density. This involves accumulating sequential LiDAR scans to form the LiDAR submap $\mathcal{S}$, and decomposing the BIM into subBIMs $\{\mathcal{B}\}$.

Given one LiDAR scan captured at the sensor frame \( \mathcal{P}_k \), with \( k \) representing the timestamp, we denote \( \mathcal{P}_1, \mathcal{P}_2, \ldots, \mathcal{P}_n \) as a sequential of LiDAR scans with consecutive timestamps. A LiDAR submap could be obtained by accumulating sequential scans with their relative pose transformations:
\begin{equation}
\mathcal{S} = \Gamma\left(\bigcup_{k=1}^n \mathbf{T_k}\mathcal{P}_k, r_v\right) ,
\end{equation}
where \( \bigcup \) denotes the union of the point sets from each LiDAR scan; $\mathbf{T_k}$ denotes the transformation from the body frame to the world frame at each timestep, which is derived from LiDAR odometry~\cite{he2023point}; \( \Gamma(\mathcal{S}, r) \) is the voxelization downsamling function, where \( r \) represents the voxel size. The parameter \( n \) is the minimum number of scans required such that the traversed distance exceeds \( d_s \) meters.

% \begin{figure}[t]
%   \centering
%   \includegraphics[width=9cm]{figs/BIMflow.pdf}
%   \caption{The decomposition process of BIM involves several steps. Initially, wall segments are extracted directly from the BIM using specialized software. Following this, a sliding window technique is applied to generate subBIMs. Within each subBIM, crucial features such as lines and corners are identified by utilizing the wall segments.}
%   \label{fig:BIMflow}
% \end{figure}

\subsubsection{BIM Preprocessing} 
\label{sec:bim_de}

We design and implement a pipeline to pre-process the BIM, focusing on extracting wall elements. The BIM model is first loaded in Revit\footnote{\href{https://www.autodesk.com/products/revit/overview/}{Revit at https://www.autodesk.com/products/revit/overview/}}. Then we retain only wall components and filter out other elements according to the category by using Rhino.Inside®.Revit\footnote{\href{https://www.rhino3d.com/inside/revit/1.0/}{Rhino.Inside.Revit at https://www.rhino3d.com/inside/revit/1.0/}}, which is a plugin that enables data exchange
between Rhino\footnote{\href{https://www.rhino3d.com/}{Rhino at https://www.rhino3d.com/}} and Revit. Walls are then directly extracted using the ``Element Curve'' scripting component within the plugin. For each floor, the associated walls are projected onto a bird's-eye view to identify corners and triangular descriptors, as illustrated in Section~\ref{sec:feature} and \ref{sec:descriptor}.

% , as detailed in Section~\ref{sec:feature}. These corners are then used to construct triangular descriptors, which are described in detail in Section~\ref{sec:descriptor}.

It is worth noting that the BIM model is not divided into sub-areas. Instead, we perform a one-shot registration between the LiDAR submap and the entire BIM model. Furthermore, the BIM preprocessing, such as corner extraction and descriptor generation, is conducted offline. Thus, during the registration stage, the precomputed descriptor database of the BIM is loaded to ensure efficient use.

% BIM data provides a digital representation of building elements, facilitating the direct extraction of semantic features such as walls using specialized tools or software, such as Revit\footnote{\href{https://www.rhino3d.com/inside/revit/1.0/}{Rhino.Inside.Revit at https://www.rhino3d.com/inside/revit/1.0/}}. After extracting these features, BIM decomposition is employed to manage the complex structure efficiently. This involves using a sliding window technique that moves along the $x$- and $y$-axes of the BIM floor plan at fixed intervals, with a predefined overlap ratio between each generated subBIM. The decomposition process is depicted in Figure~\ref{fig:BIMflow}.

% \begin{figure}[t] %!t
% \centering
% \includegraphics[width=3.5in]{figs/sub-BIM.png}
% \caption{Representation of the mechanism induced by traps on the average drain current.}
% \label{fig:subBIM}
% \end{figure}

\begin{figure*}[t]
  \centering
  \includegraphics[width=\textwidth]{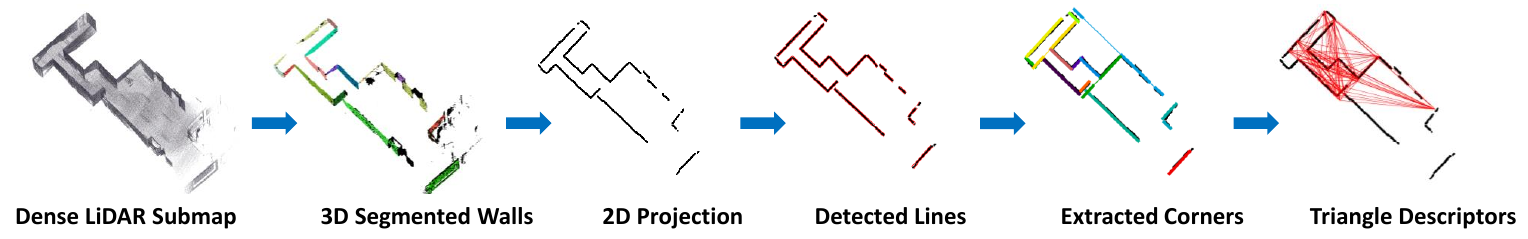}
  \caption{The workflow for extracting triangle descriptors from LiDAR submaps begins with a dense LiDAR submap. The process starts with plane segmentation to identify 3D wall structures. These segmented walls are projected onto the 2D ground plane, where lines are detected. Corner points are then extracted from the intersections of adjacent lines. Finally, triangle descriptors are computed based on the corner triplets within the submap.}
  \label{fig:extraction}
\end{figure*}

\subsection{From 3D Walls to 2D Corners}
\label{sec:feature}

In typical indoor environments, various semantic elements such as floors, columns, and walls exist~\cite{yin2023semantic}. Among these elements, the wall is a fundamental component of modern architecture. We analyze the BIM model of our university building and find that walls have the highest proportion ($>$30\%) and distribution density compared to other elements. In this study, we propose selecting walls and corners as the shared features for LiDAR and BIM data.

% However, accurately modeling walls from LiDAR data using traditional point cloud processing techniques can be time-consuming. Additionally, learning-based wall recognition methods often require additional data to support training. These factors pose challenges in achieving large-scale global registration, especially on resource-constrained platforms.

% On the other hand, the inertial measurement unit (IMU) is a gold standard for modern robotic sensing platforms. IMU can provide the gravity direction of the mapping platform, hence, the 6-degree of freedom (DoF) registration problem can be reduced to 3-DoF in the case of a planar floor. Thus, our basic idea is to project the 3D walls to 2D lines and extract corner points as features, thus generating descriptors for the downstream data association and 3-DoF pose estimation ($x-y$ coordinates and the yaw angle).

We have introduced the wall extraction in the previous BIM preprocessing. The following part mainly focuses on feature extraction for LiDAR submaps. Figure~\ref{fig:extraction} illustrates the pipeline for feature extraction from the LiDAR submap. First, plane detection is performed on a dense LiDAR submap to identify potential walls and the ground. To segment the planes, the point cloud data is discretized into voxels using an Octree structure, with the coarsest resolution denoted as \( s_v \). Each voxel contains a collection of points \( \mathbf{p}_i \), where \( i = 1, \ldots, N \). Then, the covariance matrix \( \boldsymbol{\Sigma} \) for the points within each voxel is calculated as follows:
\begin{equation}
\overline{\mathbf{p}} = \frac{1}{N} \sum_{i=1}^N \mathbf{p}_i, \quad \boldsymbol{\Sigma} = \frac{1}{N} \sum_{i=1}^N \left( \mathbf{p}_i - \overline{\mathbf{p}} \right) \left( \mathbf{p}_i - \overline{\mathbf{p}} \right)^T.
\end{equation}
To determine whether the points in a voxel belong to a plane, we analyze the eigenvalues of the covariance matrix \( \boldsymbol{\Sigma} \). Specifically, the ratio of the second smallest eigenvalue \( \lambda_2 \) to the smallest eigenvalue \( \lambda_3 \) is computed. If this ratio exceeds a predefined threshold \( \sigma_{\lambda} \), the voxel is considered to belong to a plane.

If the ratio does not meet the threshold, the voxel is further subdivided into four sub-voxels, and the plane identification process is repeated until each sub-voxel contains fewer than four points. Next, for voxels identified as belonging to a plane, neighboring planar voxels are merged if their normal directions are similar and their point-to-plane distances are small. This merging process enables the extraction of complete structures, such as walls or the ground.

Once plane segmentation is completed, the wall and ground planes are distinguished based on the angular difference between their normals and the gravity direction provided by widely used IMU. The process of extracting 2D corners from 3D walls is as follows:

\begin{enumerate}
\item Wall points are projected onto the ground plane and rasterized at a pixel scale of $s_{\mathrm{I}}$. A line segment detector based on Hough Transform is then applied to identify potential wall structures.

\item We discard 2D line segments when they are shorter than $L_{\mathrm{min}}$ to eliminate insignificant segments. The remaining 2D line segments are merged based on the distance between their endpoints and the similarity of their directions. The merged segments are then refitted using the original 2D points to accurately represent the wall geometry.

\item Finally, the refitted line segments are extended by a predefined distance to intersect and determine the corner points. To ensure the extracted corners are not too close to each other, they are refined using Non-Maximum Suppression (NMS).
\end{enumerate}

Following a similar pipeline used for the LiDAR submap, we estimate corners of the BIM from the extracted walls. This consistent approach across LiDAR submaps and BIM facilitates subsequent data association and back-end pose estimation.

\begin{figure}[t]
  \centering
  \includegraphics[width=8cm]{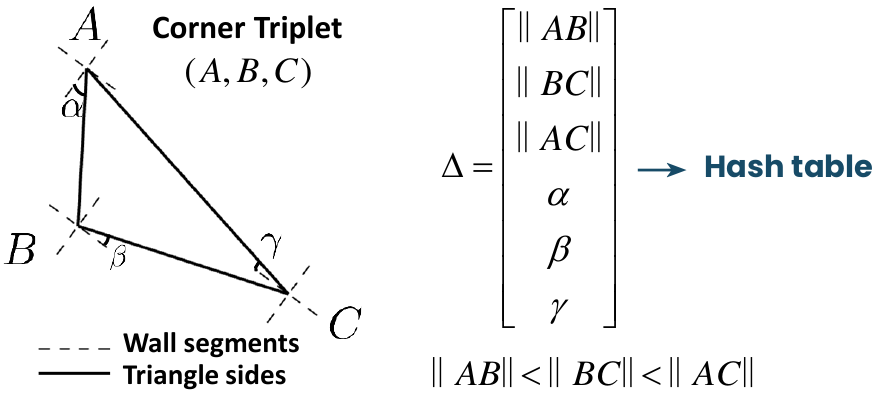}
  \caption{Triangular descriptor formulation, where each vertex in the corner triplet, denoted as $A$, $B$, and $C$, corresponds to the wall corners. The angles $\alpha$, $\beta$, and $\gamma$ represent the angles between the triangle sides and the local wall segments.}
  \label{fig:descriptor}
\end{figure}

\subsection{Triangle Descriptors}
\label{sec:descriptor}

Inspired by the STD (Stable Triangle Descriptor) \cite{yuan2023std}, we leverage both corners and wall segments to construct triangle descriptors. These descriptors are used to establish correspondences between the BIM and the LiDAR submap.

Corners extracted from the LiDAR submap and the BIM are used to form a neighborhood graph, where vertices represent corners and edges connect two corners if they are within a maximum distance of $L_{max}$. Cliques in this graph with three vertices are used to form triangle descriptors, denoted by $\Delta$. These triangles are formed by corner triplets $(A, B, C)$, as illustrated in Figure~\ref{fig:descriptor}. The attributes of a triangle descriptor $\Delta$ include:

\begin{itemize}
    \item Side lengths \(\|AB\|, \|BC\|, \|AC\|\): These are ordered such that \(\|AB\| \leq \|BC\| \leq \|AC\|\), encoding the geometric shape of the triangle.
    \item Angles \(\alpha, \beta, \gamma\): These angles are defined as the smaller angle between a triangle side and the intersecting wall segments. Specifically, $\alpha = \min(\angle(AB, l_m), \angle(AB, l_n))$ where $A$ is the intersection of wall segments $l_m$ and $l_n$. Similar definitions apply for $\beta$ and $\gamma$, associated with sides $BC$ and $AC$, respectively.
\end{itemize}

\begin{figure}[t] %!t
\centering
\includegraphics[width=9cm]{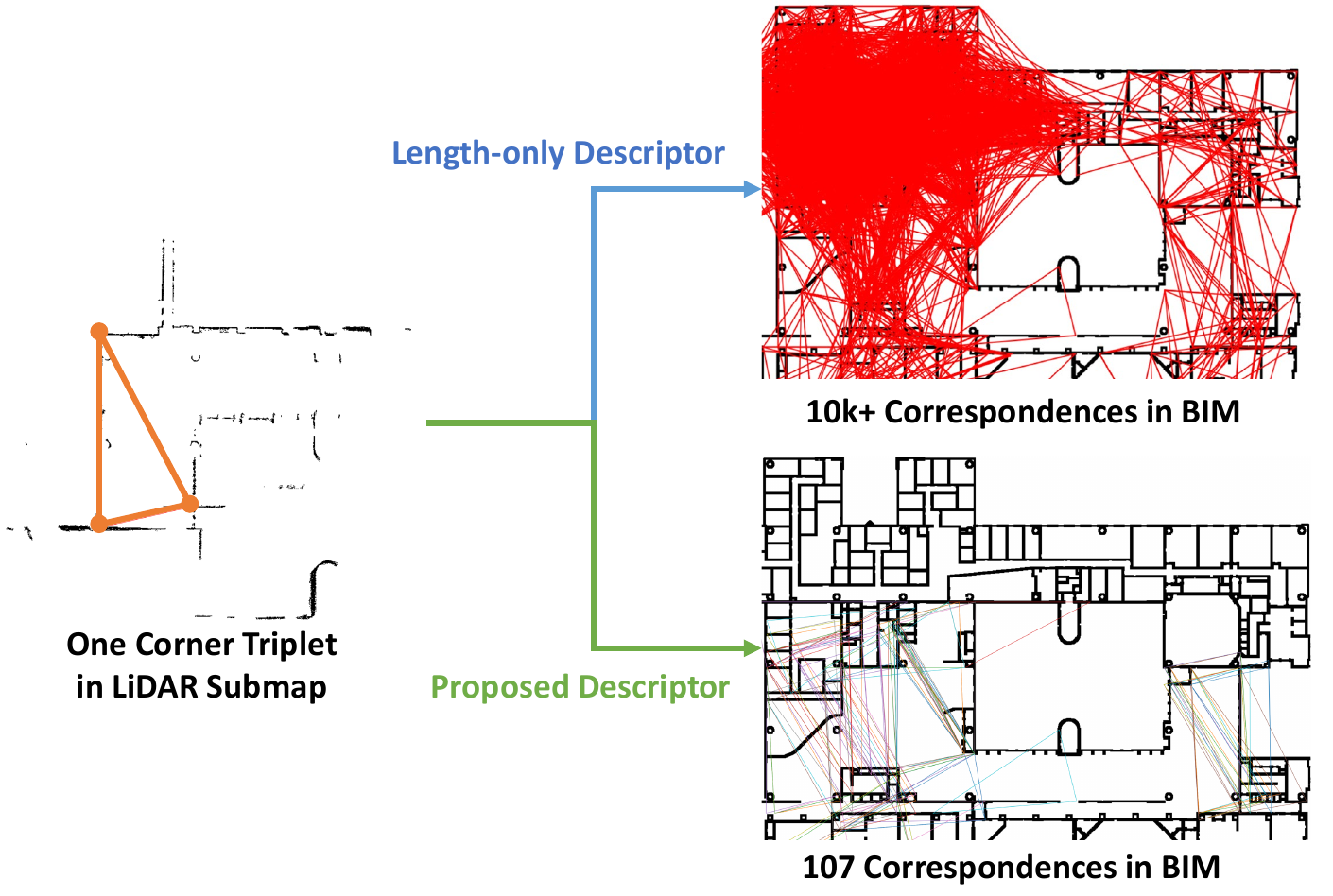}
\caption{The figure demonstrates the effectiveness of triangle descriptors in matching results, with and without angle inclusion. On the left, a highlighted corner triplet from the LiDAR submap is shown in \textcolor{orange}{orange}, which has multiple matches from the BIM using descriptor-based matching. Incorporating angle information significantly enhances the discriminative power of the triangle descriptor by encoding the local geometric structure around the triangle, which markedly reduces the false positives.}
\label{fig:orientation}
\end{figure}

These sorted side lengths and angles capture both the geometric and structural characteristics of the local environment, distinguishing this approach from the original STD descriptor \cite{yuan2023std}. Furthermore, they are invariant to rotation and translation, ensuring independence from the coordinate systems of the BIM and submaps. As shown in Figure~\ref{fig:orientation}, querying each triangle in the submap can yield multiple responses in the BIM due to repetitive architectural patterns. The inclusion of angles \(\alpha, \beta, \gamma\) helps to differentiate triangles that have identical shapes but differ in their angular relationships with adjacent wall segments, thus significantly reducing incorrect correspondences.

For efficient data management and retrieval, we construct a hash table for both LiDAR submap and the BIM. The keys are formed using the triangle descriptor $\Delta = [\|AB\|, \|BC\|, \|AC\|, \alpha, \beta, \gamma]$, with $\Delta \in \mathbf{R}^6$, and are quantized based on side length resolution $r_{\mathrm{s}}$ and angle resolution $r_{\mathrm{a}}$ as detailed in Table \ref{tab:para} and analyzed in Section \ref{exp:ab:tri}. Multiple corner triplets that yield the same hash key are stored together:
$$
\begin{aligned}
\Delta \mathcal{S} & = \{\Delta_i^s \mid \mathcal{F}(\Delta_i^s) = \{(A, B, C)_j\}_{j=1}^{K_i^s}\}_{i=1}^{N^s}, \\
\Delta \mathcal{B} & = \{\Delta_i^b \mid \mathcal{F}(\Delta_i^b) = \{(A, B, C)_j\}_{j=1}^{K_i^b}\}_{i=1}^{N^b},
\end{aligned}
$$
in which $\mathcal{F}(\Delta)$ retrieves the set of corner triplets $(A, B, C)$ corresponding to the descriptor key $\Delta$. The indices $N^s$ and $N^b$ denote the number of unique triangle descriptors in the submap and the BIM, respectively, while $K^s_i$ and $K^b_i$ represent the number of corner triplets associated with each descriptor $\Delta_i^s$ and $\Delta_i^b$.

All triangle descriptor keys and their associated corner triplets from the BIM are precomputed and stored in a hash database. During the registration stage, this database is loaded, and descriptors from the submaps are matched against the descriptors in the BIM. The matching process is performed using hash retrieval, which operates with a constant-time query complexity of $O(1)$. For each matched descriptor pair, all corresponding corner triplet candidates are exhaustively evaluated. Although this process generates an extremely large number of candidate correspondences with a high outlier ratio, it also retains a high likelihood of including the true correspondences. The next section will describe the method used to identify these true correspondences.

\section{Back-end: Transformation Voting and Verification}
\label{sec:backend}

In this section, Section~\ref{sec:hough} outlines the methodology for deriving pose transformation candidates from correspondences. This is achieved by using the Hough transform combined with a hierarchical voting mechanism. Subsequently, Section~\ref{sec:verify} introduces a verification scheme that identifies the optimal transformation candidate by employing an occupancy-aware confidence score to evaluate alignment quality, thereby ensuring reliable registration.

\subsection{Pose Hough Transformation and Hierarchical Voting}
\label{sec:hough}

The Hough transform is a robust technique in computer vision for converting raw measurements into a parameter space, where optimal parameters are identified through a hierarchical voting mechanism to effectively reject outliers.

\subsubsection{Parameter Space} 

The parameter space is based on the $\operatorname{SE}(2)$ Lie group, which represents planar rigid body transformations with 2 degrees of freedom (DOF) for translation and one for rotation (yaw). This space is discretized into voxels at resolutions defined by $r_{\mathrm{xy}}$ for x-y translation and $r_{\mathrm{yaw}}$ for yaw, as detailed in Table \ref{tab:para} and further discussed in the ablation study (Section \ref{exp:ab:voting}). For each corner triplet correspondence $\left[ (A, B, C)_s, (A, B, C)_b \right]$, transformations $(x, y, \text{yaw})$ are computed using a closed-form solution for 2D registration \cite{arun1987least}. Each transformation contributes a vote for the corresponding voxel in the parameter space.

\subsubsection{Hierarchical Voting} 

\begin{figure}[t] %!t
\centering
\includegraphics[width=\linewidth]{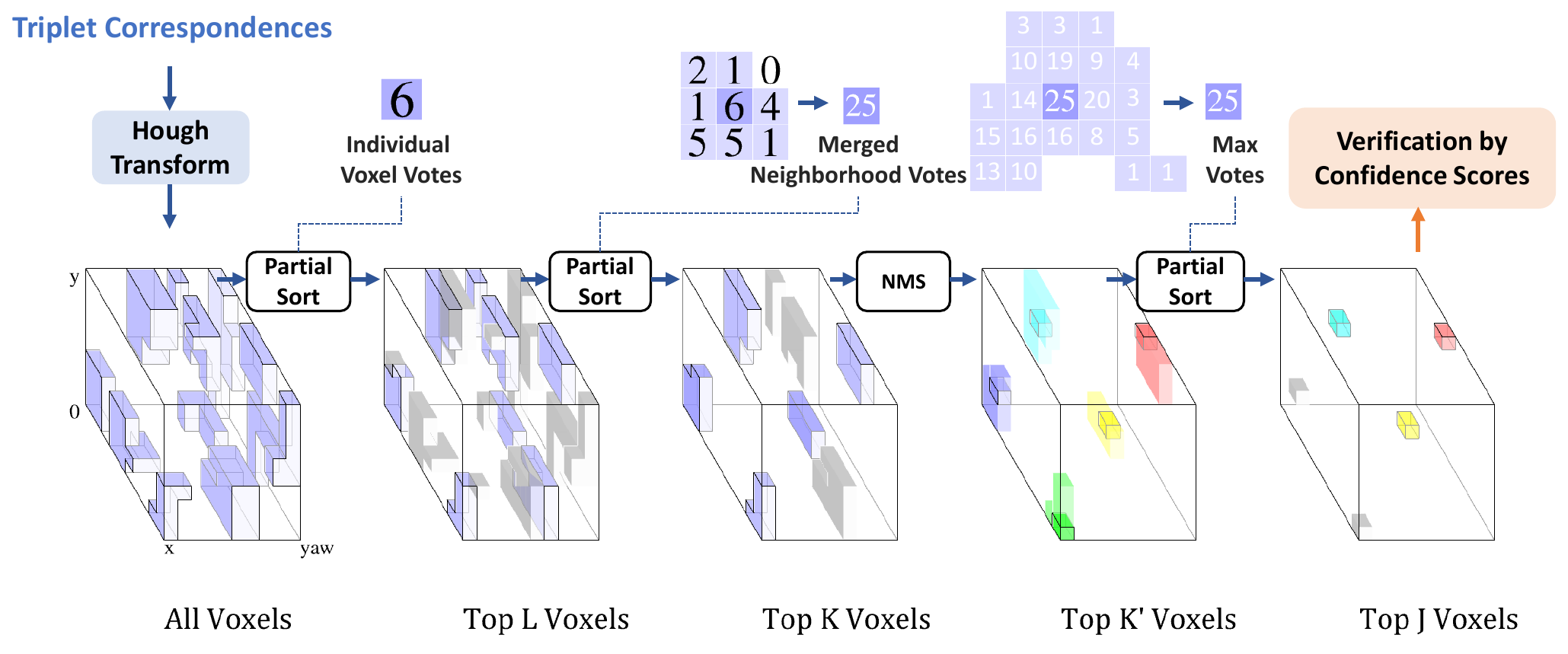}
\caption{The proposed hierarchical voting method consists of three steps: Broad Filtering, Neighborhood Refinement, and NMS with Ranking. Each step gradually reduces the number of transformation candidates, balancing computational efficiency and robustness.}
\label{fig:voting}
\end{figure}

To identify the most probable transformation candidates, we propose a hierarchical voting method that balances computational efficiency with robustness. The method comprises three steps:

\begin{enumerate}
    \item \textbf{Broad Filtering:} The top $L$ voxels are selected through partial sorting based on their individual vote counts. This computationally efficient step reduces the dataset size while preserving a pool of potential transformation candidates for further processing.
    
    \item \textbf{Neighborhood Refinement:} The $L$ selected voxels are refined to $K$ by merging votes within a $3 \times 3 \times 3$ neighborhood around each voxel. This step leverages local information to mitigate inaccuracies in corner triplet estimates, as votes may otherwise fall into adjacent voxels.

    \item \textbf{NMS and Ranking:} A region-growing clustering algorithm is applied to the $K$ refined voxels, resulting in $K'$ clusters. Non-maximum suppression (NMS) is achieved by ranking each cluster according to the maximum vote count of its constituent voxels. For each cluster, the transformation parameters are averaged to derive a representative transformation. Finally, the top $J$ clusters are selected as the most likely transformation candidates.
\end{enumerate}

This hierarchical approach progressively reduces the candidate pool ($L > K > K' > J$), achieving a balance between computational efficiency and recall. By incorporating neighborhood information, the method improves robustness and ensures reliable alignment results while maintaining manageable computational costs.

\begin{figure*}[ht]
	\centering
        \subfigure[More votes with lower confidence score]{	
        \label{fig:verify-a}
		\includegraphics[width=8.5cm]{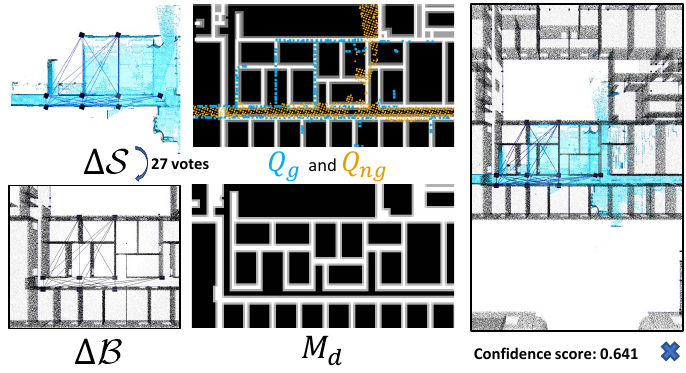}}
        \subfigure[Less votes with higher confidence score]{	
        \label{fig:verify-b}
		\includegraphics[width=8.5cm]{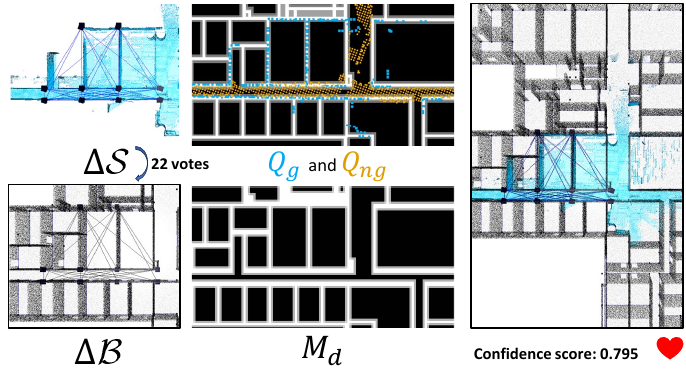}}
	\caption{Transformation verification. In each sub-figure, the first column displays the LiDAR submap and BIM along with their extracted corner triplets. The second column presents the score field \( M_d \) as well as the ground (\( Q_g \)) and non-ground (\( Q_{ng} \)) components of the transformed LiDAR submap, are presented. In the third column, the alignment using the transformation candidate is shown, along with the corresponding confidence score.  Figures \ref{fig:verify-a} and \ref{fig:verify-b} showcase two transformation candidates, where the confidence score successfully identifies the correct transformation candidate, whereas the votes do not. While both candidates align the walls and corners accurately, our proposed occupancy-aware confidence score penalizes the false candidate due to regions where the submap is free but the BIM is occupied.}
	\label{fig:verify}
\end{figure*}

\subsection{Transformation Verification}
\label{sec:verify}

Given $J$ transformation candidates, our objective is to develop a verification scheme to determine the optimal transformation which align the submap and BIM best. While corners construct the correspondences between the submap and BIM in previous sections, they don't capture enough geometric structure for alignment quality evaluation. As illustrated in Figure~\ref{fig:verify}, both transformation candidates can align the walls and corners, with the false candidate even receiving more votes from the matched corner triplets. To address this issue, we utilize the original submap point cloud to compute an occupancy-aware score. This score incorporates original and richer geometric information, offering a more robust evaluation of alignment quality to reliably identify the optimal transformation.

Denote the BIM wall point cloud as \( P_\text{B} \), and the ground and non-ground planar components of the transformed LiDAR submap as \( Q_g \) and \( Q_{ng} \), respectively. An occupancy-aware confidence score is proposed to evaluate a transformation candidate by assessing the alignment quality between the BIM and the transformed submap using the given transformation.

Specifically, a rasterization function \( \phi(\cdot, s_r) \) is used to map \( P_\text{B} \) to a 2D grid \( M \) with a resolution of \( s_r \), where occupied cells are set to 1 and free cells to 0:
\begin{equation}
M = \phi(P_\text{B}, s_r).
\end{equation}
\par To account for as-built deviations, the grid \( M \) is dilated using a kernel of size \( k_d \), producing the score field \( M_d \). The value of each cell in \( M_d \) decreases linearly with the distance to the nearest originally occupied cell, ranging from 1 at the nearest cell to \( \frac{1}{k_d} \) at the kernel's edge. This dilation introduces gradual transitions in cell values, enhancing robustness to noise and inaccuracies.

\paragraph{Award Component}  
The method rewards points for regions where the BIM wall point cloud and the actual construction \( Q_{ng} \) match. The award score \( s_a \) is calculated as:

\begin{equation}
s_a = \sum_{\mathbf{q} \in Q_{ng}} M_{d}(\phi(\mathbf{q}, s_r)),
\end{equation}
where \( M_{d}(\phi(\mathbf{q}, s_r)) \) denotes the value of the cell corresponding to the rasterized point \( \phi(\mathbf{q}, s_r) \). We do not consider the situation where both the submap and the BIM are free, as this is less important than the occupied cases and may bias the score if included.

\paragraph{Penalty Component}  
Structures included in the BIM model are expected to be reflected in the actual construction. To penalize areas where the BIM is occupied but the submap is free, the penalty score \( s_p \) is computed as:
\begin{equation}
s_p = \sum_{\mathbf{q} \in Q_g} M_d(\phi(\mathbf{q}, s_r)).
\end{equation}
\par We do not consider the situation where the submap is occupied but the BIM is free, as real construction environments typically include additional elements such as lockers and furniture, which is a normal occurrence, especially when the BIM is outdated.

\paragraph{Final Confidence Score}  
The final confidence score is defined as:
\begin{equation}
\text{score} = \frac{s_a - \lambda \cdot s_p}{|Q_{ng}|},
\end{equation}
where \( \lambda \) is a hyperparameter that balances the contributions of the award and penalty terms. The score is normalized by dividing by \( |Q_{ng}| \), the number of non-ground points in the transformed LiDAR submap, ensuring robustness to submaps of varying sizes. In the ideal case, when all points in \( Q_{ng} \) match the BIM and there is no penalty from ground points, the score reaches its maximum value of 1.

This confidence score provides a quantitative and normalized measure of how effectively the transformation aligns the submap with the BIM. Furthermore, it can be utilized by users to assess the reliability of the registration, as demonstrated in Experiment~\ref{exp:conf}. 

\section{Experiments}
\label{sec:experiments}

This section is structured as follows: In Section~\ref{exp:setup}, we introduce the dataset, evaluation metrics, and implementation details of our proposed method, as well as those of the methods against which we compare. Section~\ref{exp:recall} presents comparative experiments that assess the registration recall of our method relative to other advanced methods. In Section~\ref{exp:ab}, we explore how different parameters affect the performance of our proposed algorithm through an ablation study. In Section~\ref{exp:conf}, we demonstrate the ability of our proposed occupancy-aware score to assess the reliability of the registration. Section~\ref{exp:case} provides visualizations of our algorithm applied in real-world scenarios and discusses cases where the algorithm failed.

\begin{figure}[t] %!t
\centering
\includegraphics[width=8cm]{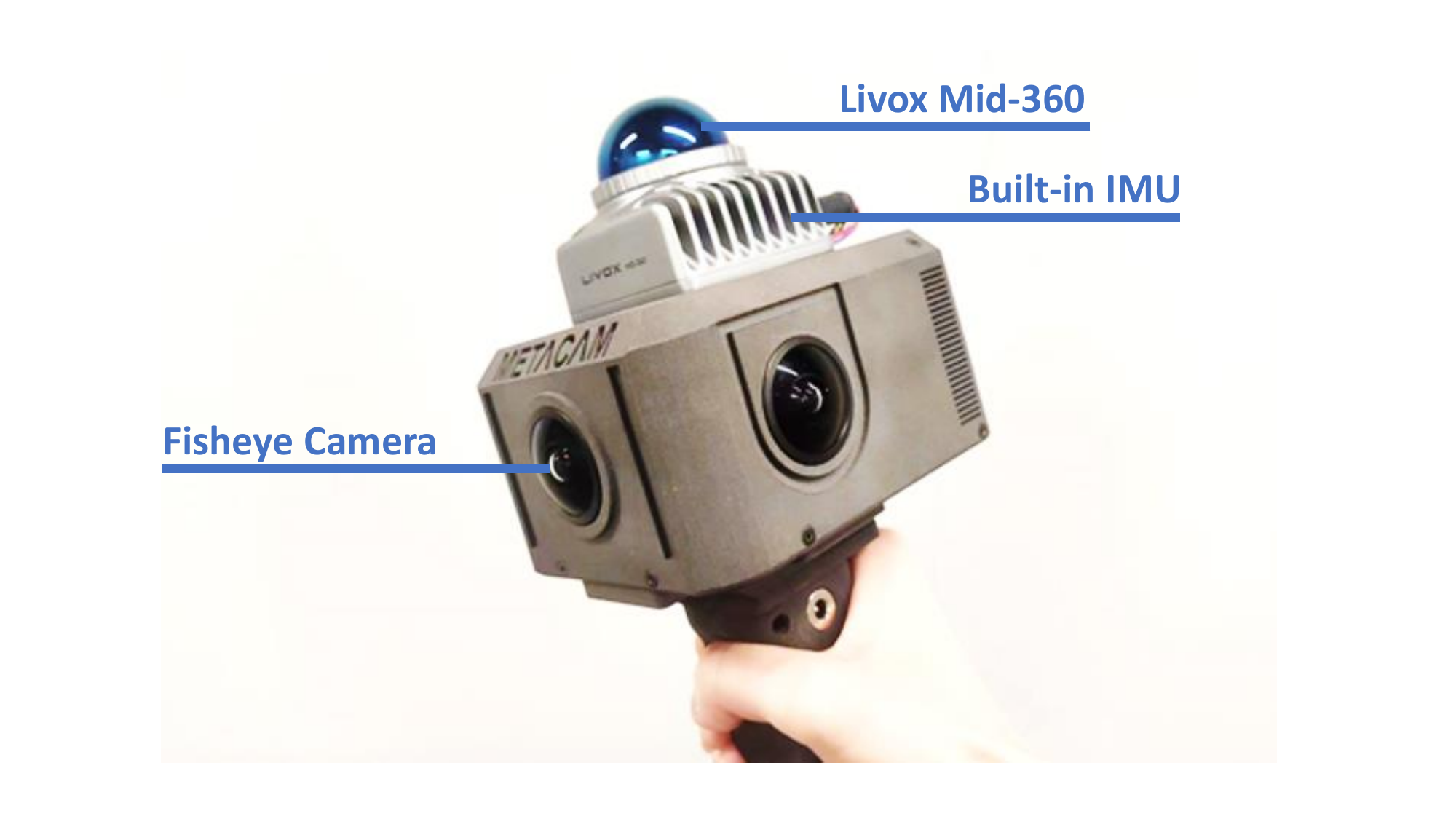}
\caption{The handheld sensor suite: a Livox Mid-360 LiDAR scanner, fisheye cameras for wide-angle imaging, and a built-in IMU.}
\label{fig:sensor}
\end{figure}

\subsection{Experimental Setup}
\label{exp:setup}

\subsubsection{LiBIM-UST Dataset}

% Please add the following required packages to your document preamble:
% \usepackage{multirow}
% \usepackage{graphicx}
\begin{table}[!t]
\centering
\caption{LiBIM-UST Dataset Information}
\label{table: seq}
\resizebox{\columnwidth}{!}{%
\begin{tabular}{c|c|c|c|c}
\hline
\hline
LiDAR Type & Sequence & Total Distance (m) & Submaps (15 m) & Submaps (30 m) \\ \hline
Ouster-128 & Building day & 402.14 & 169 & - \\ \hline
\multirow{5}{*}{Mid-360} & 1f-office & 318.24 & 118 & 111 \\
 & 2f-office & 533.78 & 193 & 171 \\
 & 3f-office & 187.55 & 44 & 36 \\
 & 4f-office & 191.52 & 76 & 69 \\
 & 5f-office & 123.63 & 27 & 19 \\ 
 \hline
 \hline
\end{tabular}%
}
\end{table}

We developed the LiBIM-UST dataset using the publicly available FusionPortable dataset~\cite{jiao2022fusionportable} and self-collected data for evaluation. The details are provided in Table~\ref{table: seq}, where \( d_s \), described in Section~\ref{sec:submap}, is set to different values for submaps of varying sizes. The sensor configuration for the self-collected data is illustrated in Figure~\ref{fig:sensor}. This dataset includes the BIM of the main academic building at the Hong Kong University of Science and Technology (HKUST). The BIM model is part of the digital twin project at HKUST, which was initiated in 2019. 

The model was developed using Autodesk Revit and is stored in the .RVT file format with a Level of Detail (LoD) of 300~\cite{chen2022building}. It was created based on archival 2D drawings from 1988, where the interior was primarily modeled from floor plans, and the exterior appearance was reconstructed using site photos and point cloud data. However, the model lacks textural information and contains several discrepancies when compared to the actual built environment as of 2023.

% \begin{figure}[t] %!t
% \centering
% \includegraphics[width=3.5in]{figs/dataset.png}
% \caption{Representation of the mechanism induced by traps on the average drain current.}
% \label{fig:subBIM}
% \end{figure}

\subsubsection{Evaluation Metric}

To evaluate the effectiveness and efficiency of the registration algorithms, we use registration recall and average time consumption, respectively. The success of registration, or true positive, is determined based on the transformation estimation $(\mathbf{R}_{est}, \mathbf{t}_{est})$ compared to the ground truth $(\mathbf{R}_{gt}, \mathbf{t}_{gt})$. A registration is considered successful if:
\begin{equation}
\begin{aligned}
    \arccos \left(\frac{\operatorname{tr}(\mathbf{R}_{est}^{T} \mathbf{R}_{gt}) - 1}{2}\right) < 5^\circ \\
    \| \mathbf{R}_{est}^{T}(\mathbf{t}_{gt} - \mathbf{t}_{est}) \|_2 < 3\text{m}
\end{aligned}
\end{equation}
\par These thresholds $(5^\circ, 3\text{m})$ are chosen based on common practice, allowing for successful local refinement methods like ICP and NDT to follow.

\subsubsection{Comparisons and Implementation}
\label{exp:details}

Given that LiDAR to BIM global registration is a novel task in cross-modal registration with limited open-source work and benchmarks, we selected representative methods from both image and point cloud registration domains for comparison.

\textbf{Image Registration:} For image registration, intermediate images from the line extraction process (Section~\ref{sec:feature}) are used. We implemented the SIFT~\cite{lowe1999object} and a combination of Superpoint~\cite{detone2018superpoint} with LightGlue~\cite{lindenberger2023lightglue}, representing traditional and deep learning-based methods, respectively. RANSAC, with up to $1 \times 10^8$ iterations, was used for pose estimation.

\textbf{Point Cloud Registration:} We employed three common global point cloud registration methods: FPFH~\cite{rusu2009fast} combined with TEASER++~\cite{yang2020teaser}, GeoTransformer~\cite{qin2022geotransformer}, an advanced deep learning based registration method, and 3D-BBS~\cite{aoki20243d}, a branch-and-bound based method. For FPFH, we set the normal search radius to 1 m and the feature search radius to 2.5 m. For TEASER++, the noise bound is set at 0.4 m. For GeoTransformer, we retrain it on our datasets. For 3D-BBS, the parameter $\mathrm{min\_level\_res}$ is set to 0.2 m and the $\mathrm{max\_level}$ is set to 6.

\textbf{Our Method:} Detailed parameters of our method are listed in Table~\ref{tab:para}. The BIM preprocessing and the Hash table configuration were predetermined. The experiments were conducted on an Intel i7-13700K CPU and an Nvidia RTX4080 graphics card.

% Please add the following required packages to your document preamble:
% \usepackage{multirow}
\begin{table}[t]
\centering
\caption{Default Parameters of The Proposed Method}
\label{tab:para}
\begin{tabular}{cc|c}
\hline
\hline
\multicolumn{2}{c|}{Parameters} & Value \\ \hline
\multicolumn{1}{c|}{\multirow{4}{*}{Wall Extraction}} & $r_{v}$ & 0.8 m \\
\multicolumn{1}{c|}{} & $\sigma_{\lambda}$ & 10 \\
\multicolumn{1}{c|}{} & $s_{\mathrm{I}}$ & 60 pixels / m \\
\multicolumn{1}{c|}{} & $L_{\mathrm{min}}$ & 30 pixels \\ \hline
\multicolumn{1}{c|}{\multirow{3}{*}{Triangle Descriptor}} & $r_{\mathrm{s}}$ & 0.5 m \\
\multicolumn{1}{c|}{} & $r_{\mathrm{a}}$ & 3.0$^{\circ}$ \\
\multicolumn{1}{c|}{} & $L_{\mathrm{max}}$ & 30.0 m / 40.0 m \\ \hline
\multicolumn{1}{c|}{\multirow{3}{*}{\begin{tabular}[c]{@{}c@{}}Hough Transform\\ and Voting\end{tabular}}} & $r_{\mathrm{xy}}$ & 0.15 m \\
\multicolumn{1}{c|}{} & $r_{\mathrm{yaw}}$ & 1.0$^{\circ}$ \\
\multicolumn{1}{c|}{} & L / K / J &  10000 / 5000 / 1500 \\ 
\hline
\hline
\end{tabular}
\end{table}

% Please add the following required packages to your document preamble:
% \usepackage{multirow}
% \usepackage{graphicx}
\begin{table*}[]
\centering
\caption{Global Registration Evaluation on LiBIM-UST Dataset}
\label{tab:benchmark}
\resizebox{\textwidth}{!}{%
\begin{tabular}{c|c|c|ccccccccccc}
\hline
\hline
 & \multirow{3}{*}{Front-end} & \multirow{3}{*}{Back-end} & \multicolumn{11}{c}{Recall (\%)} \\ \cline{4-14} 
 &  &  & \multicolumn{1}{c|}{BuildingDay} & \multicolumn{2}{c|}{1F} & \multicolumn{2}{c|}{2F} & \multicolumn{2}{c|}{3F} & \multicolumn{2}{c|}{4F} & \multicolumn{2}{c}{5F} \\ \cline{4-14} 
 &  &  & \multicolumn{1}{c|}{$d_s$=15m} & $d_s$=15m & \multicolumn{1}{c|}{$d_s$=30m} & $d_s$=15m & \multicolumn{1}{c|}{$d_s$=30m} & $d_s$=15m & \multicolumn{1}{c|}{$d_s$=30m} & $d_s$=15m & \multicolumn{1}{c|}{$d_s$=30m} & $d_s$=15m & $d_s$=30m \\ \hline
\multirow{2}{*}{\begin{tabular}[c]{@{}c@{}}Image Matching-\\ based Method\end{tabular}} & SIFT & RANSAC & \multicolumn{1}{c|}{0} & 0 & \multicolumn{1}{c|}{0} & 0 & \multicolumn{1}{c|}{0} & 0 & \multicolumn{1}{c|}{0} & 0 & \multicolumn{1}{c|}{0} & 0 & 0 \\
 & Superpoint+Lightglue & RANSAC & \multicolumn{1}{c|}{0} & 0 & \multicolumn{1}{c|}{0} & 0 & \multicolumn{1}{c|}{0} & 0 & \multicolumn{1}{c|}{0} & 0 & \multicolumn{1}{c|}{0} & 0 & 0 \\ \hline
\multirow{3}{*}{\begin{tabular}[c]{@{}c@{}}Point Cloud-\\ based Method\end{tabular}} & FPFH & TEASER++ & \multicolumn{1}{c|}{0} & 0 & \multicolumn{1}{c|}{0} & 1.04 & \multicolumn{1}{c|}{0} & 0 & \multicolumn{1}{c|}{0} & 1.32 & \multicolumn{1}{c|}{0} & 0 & 0 \\
 & GeoTransformer & Weighted SVD & \multicolumn{1}{c|}{0} & 6.67 & \multicolumn{1}{c|}{0} & 0 & \multicolumn{1}{c|}{0} & 7.69 & \multicolumn{1}{c|}{0} & 0 & \multicolumn{1}{c|}{0} & 0 & 0 \\
 & Voxel Map & 3DBBS & \multicolumn{1}{c|}{86.39} & 41.52 & \multicolumn{1}{c|}{47.74} & 75.65 & \multicolumn{1}{c|}{94.74} & \textbf{100.0} & \multicolumn{1}{c|}{\textbf{100.0}} & 67.10 & \multicolumn{1}{c|}{94.20} & \textbf{100.0} & \textbf{100.0} \\ \hline
Ours & Triangle Descriptor & Hough Voting & \multicolumn{1}{c|}{\textbf{98.81}} & \textbf{64.40} & \multicolumn{1}{c|}{\textbf{89.18}} & \textbf{77.20} & \multicolumn{1}{c|}{\textbf{98.83}} & 97.72 & \multicolumn{1}{c|}{\textbf{100.0}} & \textbf{88.15} & \multicolumn{1}{c|}{\textbf{100.0}} & \textbf{100.0} & \textbf{100.0} \\ \hline
 & \multirow{3}{*}{Front-end} & \multirow{3}{*}{Back-end} & \multicolumn{11}{c}{Average Time (ms)} \\ \cline{4-14} 
 &  &  & \multicolumn{1}{c|}{BuildingDay} & \multicolumn{2}{c|}{1F} & \multicolumn{2}{c|}{2F} & \multicolumn{2}{c|}{3F} & \multicolumn{2}{c|}{4F} & \multicolumn{2}{c}{5F} \\ \cline{4-14} 
 &  &  & \multicolumn{1}{c|}{$d_s$=15m} & $d_s$=15m & \multicolumn{1}{c|}{$d_s$=30m} & $d_s$=15m & \multicolumn{1}{c|}{$d_s$=30m} & $d_s$=15m & \multicolumn{1}{c|}{$d_s$=30m} & $d_s$=15m & \multicolumn{1}{c|}{$d_s$=30m} & $d_s$=15m & $d_s$=30m \\ \hline
\multirow{2}{*}{\begin{tabular}[c]{@{}c@{}}Image Matching-\\ based Method\end{tabular}} & SIFT & RANSAC & \multicolumn{1}{c|}{6786.33} & 10670.65 & \multicolumn{1}{c|}{8946.49} & 12935.78 & \multicolumn{1}{c|}{13034.09} & 7083.24 & \multicolumn{1}{c|}{5618.32} & 12683.18 & \multicolumn{1}{c|}{14477.53} & 8283.46 & 5092.87 \\
 & Superpoint+Lightglue & RANSAC & \multicolumn{1}{c|}{499.93} & 859.68 & \multicolumn{1}{c|}{1209.79} & 500.43 & \multicolumn{1}{c|}{794.47} & 302.09 & \multicolumn{1}{c|}{392.47} & 650.38 & \multicolumn{1}{c|}{791.70} & 317.18 & 486.24 \\ \hline
\multirow{3}{*}{\begin{tabular}[c]{@{}c@{}}Point Cloud-\\ based Method\end{tabular}} & FPFH & TEASER++ & \multicolumn{1}{c|}{25694.49} & 4169.82 & \multicolumn{1}{c|}{4799.47} & 473.214 & \multicolumn{1}{c|}{921.16} & 455.51 & \multicolumn{1}{c|}{818.19} & 517.70 & \multicolumn{1}{c|}{666.60} & 601.38 & 717.67 \\
 & GeoTransformer & Weighted SVD & \multicolumn{1}{c|}{441.3} & 492.7 & \multicolumn{1}{c|}{275.0} & 378.9 & \multicolumn{1}{c|}{397.3} & 427.5 & \multicolumn{1}{c|}{366.7} & 415.7 & \multicolumn{1}{c|}{432.2} & 467.1 & 422.9 \\
 & Voxel Map & 3DBBS & \multicolumn{1}{c|}{6603.00} & 2850.90 & \multicolumn{1}{c|}{5284.33} & 2083.00 & \multicolumn{1}{c|}{2862.00} & 821.72 & \multicolumn{1}{c|}{1408.35} & 3658.24 & \multicolumn{1}{c|}{4158.97} & 1373.10 & 1674.25 \\ \hline
Ours & Triangle Descriptor & Hough Voting & \multicolumn{1}{c|}{\textbf{121.34}} & \textbf{36.72} & \multicolumn{1}{c|}{\textbf{95.14}} & \textbf{54.96} & \multicolumn{1}{c|}{\textbf{170.12}} & \textbf{77.42} & \multicolumn{1}{c|}{\textbf{369.60}} & \textbf{43.30} & \multicolumn{1}{c|}{\textbf{152.83}} & \textbf{68.79} & \textbf{268.59} \\ 
\hline
\hline
\end{tabular}%
}
\end{table*}

\subsection{Performance on Registration Recall}
\label{exp:recall}

The registration recall results on the LiBIM-UST dataset are summarized in Table \ref{tab:benchmark}. This table compares the performance of image-based registration, point cloud-based registration, and our method.

Methods relying on local descriptors, such as SIFT~\cite{lowe1999object}, Superpoint~\cite{detone2018superpoint}, and FPFH~\cite{rusu2009fast}, all recorded a 0\% registration recall. Despite attempts to optimize these methods by adjusting descriptor parameters and increasing the number of correspondences (typical counts were 400 for SIFT and Superpoint, and 4000 for FPFH), none were successful. Similarly, the learning-based Geotransformer~\cite{qin2022geotransformer} suffers from the cross-modality challenges and result in an extremely low registration recall. The collective failure of local descriptor-based methods highlights the ineffectiveness of this approach for LiDAR and BIM registration. This is because these methods typically rely on discriminative local features, such as normals and gradients, which cannot be adequately provided by BIM. BIM structures predominantly consist of planar surfaces, where normals are mostly parallel or perpendicular, offering limited discriminative power. In contrast, our proposed triangle descriptor avoids relying on these constrained local features and instead harnesses more distinctive and shared structural information, enabling more robust data association.

The 3D-BBS method~\cite{aoki20243d} showed potential for LiDAR-to-BIM registration by searching through the transformation space to align submaps closely with the BIM. However, it does not account for discrepancies between as-designed and as-built conditions, leading to lower recall on most sequences. Additionally, since it searches the entire transformation space, 3D-BBS is computationally expensive and time-consuming. In contrast, our triangle descriptor-based matching drastically narrows the search scope. Furthermore, we introduce a hierarchical voting mechanism to accelerate the search process, significantly reducing computation time. The occupancy-aware score design further improves robustness to construction discrepancies, enabling a more reliable evaluation of alignment quality for each transformation candidate.

\begin{figure}[ht]
  \centering
  \subfigure[]{
    \includegraphics[width=0.45\linewidth]{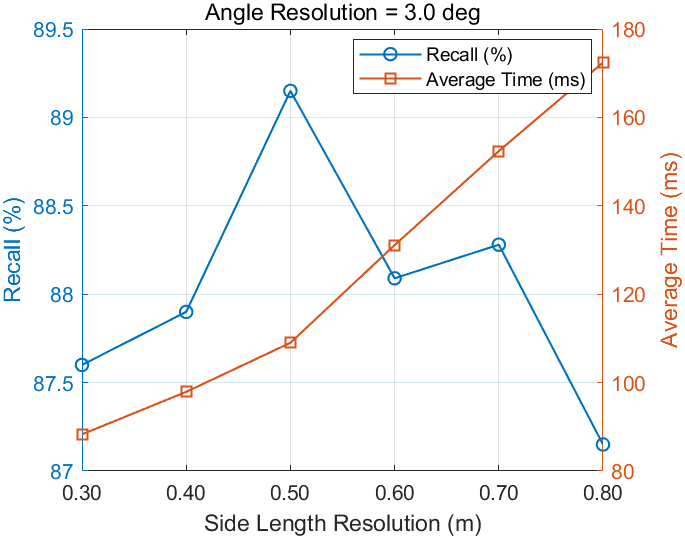}
    \label{fig:ab_study:a}
  }
  \hspace{-0.2cm}
  \vspace{-0.2cm}
  \subfigure[]{
    \includegraphics[width=0.45\linewidth]{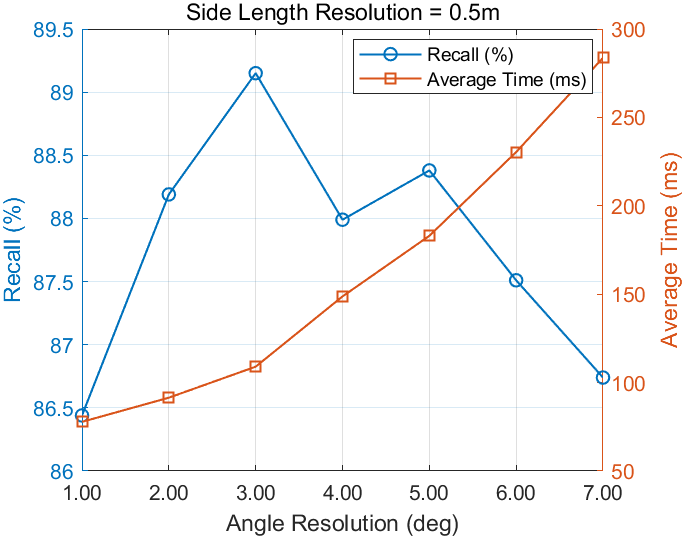}
    \label{fig:ab_study:b}
  }
  \subfigure[]{
    \includegraphics[width=0.45\linewidth]{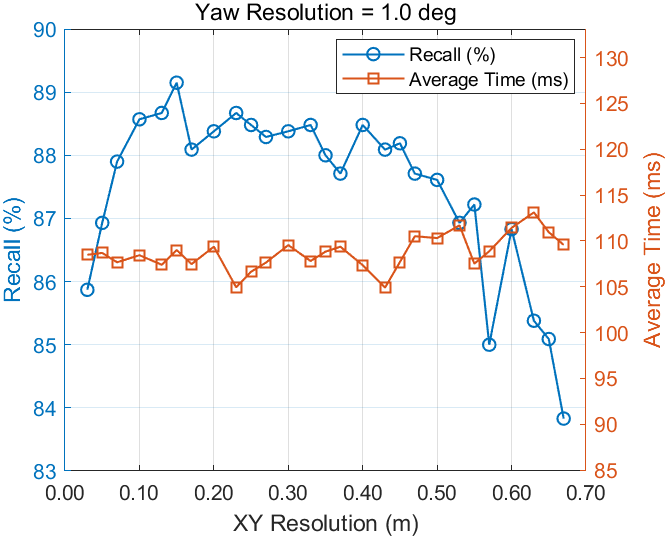}
    \label{fig:ab_study:c}
  }
  \hspace{-0.2cm}
  \vspace{-0.2cm}
  \subfigure[]{
    \includegraphics[width=0.45\linewidth]{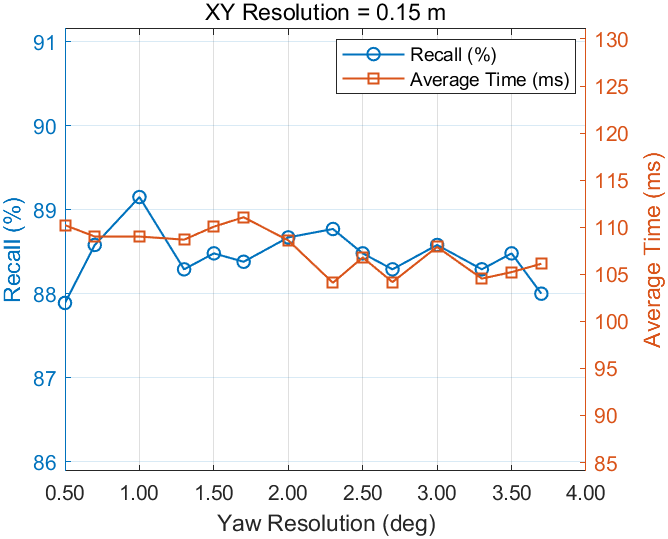}
    \label{fig:ab_study:d}
  }
  \caption{(a) and (b): The registration recall and average time with varying side length resolution \( r_{\mathrm{s}} \) and angle resolution \( r_{\mathrm{a}} \) for building triangle descriptors. (c) and (d): The registration recall and average time with varying translation resolution \( r_{\mathrm{xy}} \) and yaw resolution \( r_{\mathrm{yaw}} \) in the Hough transform.}
  \label{fig:ab_study}
\end{figure}
% Please add the following required packages to your document preamble:
% \usepackage{multirow}
% \usepackage{graphicx}
\begin{table}[h]
\centering
\caption{ABLATION STUDY ON LiBIM-UST DATASET}
\label{tab:ab_study}
\resizebox{\columnwidth}{!}{%
\begin{tabular}{c|c|c|c}
\hline
\hline
Variable & Value & Registration Recall(\%) & Average Time(ms) \\ \hline
\multirow{3}{*}{\begin{tabular}[c]{@{}c@{}}Triangle\\ Descriptor $r_{\mathrm{l}}$\end{tabular}} & 0.3m & 87.60 & 88.29 \\
 & 0.5m & \textbf{89.15} & \textbf{109.01} \\
 & 0.7m & 88.28 & 152.38 \\ \hline
\multirow{3}{*}{\begin{tabular}[c]{@{}c@{}}Triangle \\ Descriptor  $r_{\mathrm{a}}$\end{tabular}} & 1$^{\circ}$ & 86.44 & 77.80 \\
 & 3$^{\circ}$ & \textbf{89.15} & \textbf{109.01} \\
 & 5$^{\circ}$ & 88.38 & 183.02 \\ \hline
\multirow{3}{*}{\begin{tabular}[c]{@{}c@{}}Hough\\ Transform $r_{\mathrm{xy}}$\end{tabular}} & 0.10m & 88.57 & \textbf{108.44} \\
 & 0.15m & \textbf{89.15} & 109.01 \\
 & 0.20m & 88.38 & 109.37 \\ \hline
\multirow{3}{*}{\begin{tabular}[c]{@{}c@{}}Hough\\ Transform $r_{\mathrm{yaw}}$\end{tabular}} & 0.5$^{\circ}$ & 87.89 & 110.22 \\
 & 1.0$^{\circ}$ & \textbf{89.15} & \textbf{109.01} \\
 & 1.5$^{\circ}$ & 88.48 & 110.06 \\ \hline
\multirow{5}{*}{Hierachical Voting} & Vanilla & 38.92 & \textbf{75.99} \\
 & Layer 1 & \textbf{89.64} & 134.26 \\
 & Layer 2 & 89.16 & 166.94 \\
 & all & 89.15 & 109.01 \\ \hline
\multirow{4}{*}{\begin{tabular}[c]{@{}c@{}}Transformation\\ Estimator\end{tabular}} 
& SVD & 0 & \textbf{55.45} \\
& RANSAC & 8.03 & 24027.81 \\
& TEASER++ & 25.07 & 13851.75 \\
& Hough Voting & \textbf{89.15} & 109.01 \\ 
\hline
\hline
\end{tabular}%
}
\end{table}

\subsection{Ablation Study}
\label{exp:ab}

In this section, we explore the effects of several key parameters on the overall performance of our registration system. The parameters under evaluation include the building of the triangle descriptor, the Hough transform, hierarchical voting, and the transformation estimator. All sequences are considered for evaluation.

\subsubsection{Triangle descriptor}
\label{exp:ab:tri}

The side length resolution \( r_{\mathrm{s}} \) and angle resolution \( r_{\mathrm{a}} \) account for inaccuracies in corner extraction from the submaps and the differences between the as-designed BIM and the as-built environment, as the triangle descriptors do not need to be identical; they can still be matched as long as they fall into the same hash key.
 
As shown in Figure~\ref{fig:ab_study:a}-\ref{fig:ab_study:b} and Table~\ref{tab:ab_study}, both excessively high and low resolutions adversely affect registration recall. Resolutions that are excessively fine prevent correct corner triplet correspondences from being grouped together, leading to an increase in false negatives. Conversely, excessively coarse resolutions may group incorrect correspondences together, increasing false positives and introducing incorrect transformation candidates during the voting process. This not only complicates the transformation verification process but also significantly increases computation time due to the higher number of corner triplets sharing the same descriptor, especially with an all-to-all matching strategy (see Section~\ref{sec:descriptor}).

\subsubsection{Hough Transform}
\label{exp:ab:voting}

The use of \( r_{\mathrm{xy}} \) is analogous to \( r_{\mathrm{s}} \) and \( r_{\mathrm{a}} \), as both discretize the space into voxels to group similar values together. Resolutions that are too fine scatter the votes, preventing them from converging around the true transformation. Conversely, excessively coarse resolutions reduce the precision of the estimated transformations and may introduce false positives during transformation verification. 

However, the Hough transform is less sensitive to resolution changes than the triangle descriptor construction, as the hierarchical voting mechanism effectively filters out false candidates, as shown in Figure~\ref{fig:ab_study:c} and \ref{fig:ab_study:d}. Additionally, the yaw resolution is less critical than the translation resolution, as our method primarily focuses on wall alignment, where yaw candidates are typically multiples of 90 degrees.

\subsubsection{Hierarchical Voting}

To compare the proposed hierarchical voting method with the original voting method, we evaluate our hierarchical scheme by retaining only the first \( n \) layers, such as Layer 1 and Layer 2, where \( n = 1 \) and \( n = 2 \), with the default parameters \( L = 10000 \) and \( K = 5000 \), respectively. Additionally, we implement the original voting method, referred to as vanilla voting, which selects the transformation with the highest number of votes as the final result.

As shown in Table~\ref{tab:ab_study}, Layer 1 achieves the best recall performance because it retains the largest number of transformation candidates. However, its runtime is the longest due to the computational burden of the verification step. On the other hand, vanilla voting exhibits the worst performance, as relying solely on the vote count as a criterion is limited by the insufficient information available from corner matching, as discussed in Section~\ref{sec:verify}. Conversely, the proposed hierarchical voting method achieves an optimal balance between efficiency and robustness.

\subsubsection{Back-end Estimation}
\label{exp:ab:backend}

We compare the proposed transformation estimator with other alternative estimators, such as ordinary least squares using SVD \cite{arun1987least}, the robust estimator TEASER++ \cite{yang2020teaser}, and RANSAC \cite{ransacfamily}. As shown in Table~\ref{tab:ab_study}, the proposed method achieves the best performance in terms of registration recall and average time consumption. SVD is highly sensitive to outliers and noise, resulting in a lower recall rate. Although TEASER++ is robust to outliers, it provides only a single transformation candidate, which may overlook the true transformation. RANSAC employs a hypothesis-and-verification strategy; however, the score of each candidate relies solely on consistent corner correspondences, which may be insufficient to comprehensively evaluate the transformation quality.

In terms of computational complexity, the proposed method achieves $O(N)$ complexity for the Hough transform and $O(N \log(N))$ for hierarchical voting. In comparison, alternative methods such as TEASER++ \cite{yang2020teaser} have a worst-case complexity of $O(n^2)$, where $N$ and $n$ denote the number of corner triplet correspondences and the number of corner correspondences, respectively, with $n < N \ll n^2$. Similarly, RANSAC-based methods \cite{ransacfamily} can exhibit exponential complexity under high outlier ratios. Furthermore, since calculating transformations from corner triplet correspondences and evaluating their confidence scores are independent processes, they can be executed in parallel. This parallelization significantly accelerates real-world implementation.

% The proposed method offers improved computational efficiency, making it well-suited for large-scale and noisy registration tasks while maintaining robustness in the presence of outliers.

\begin{figure}[t] %!t
\centering
\includegraphics[width=0.9\linewidth]{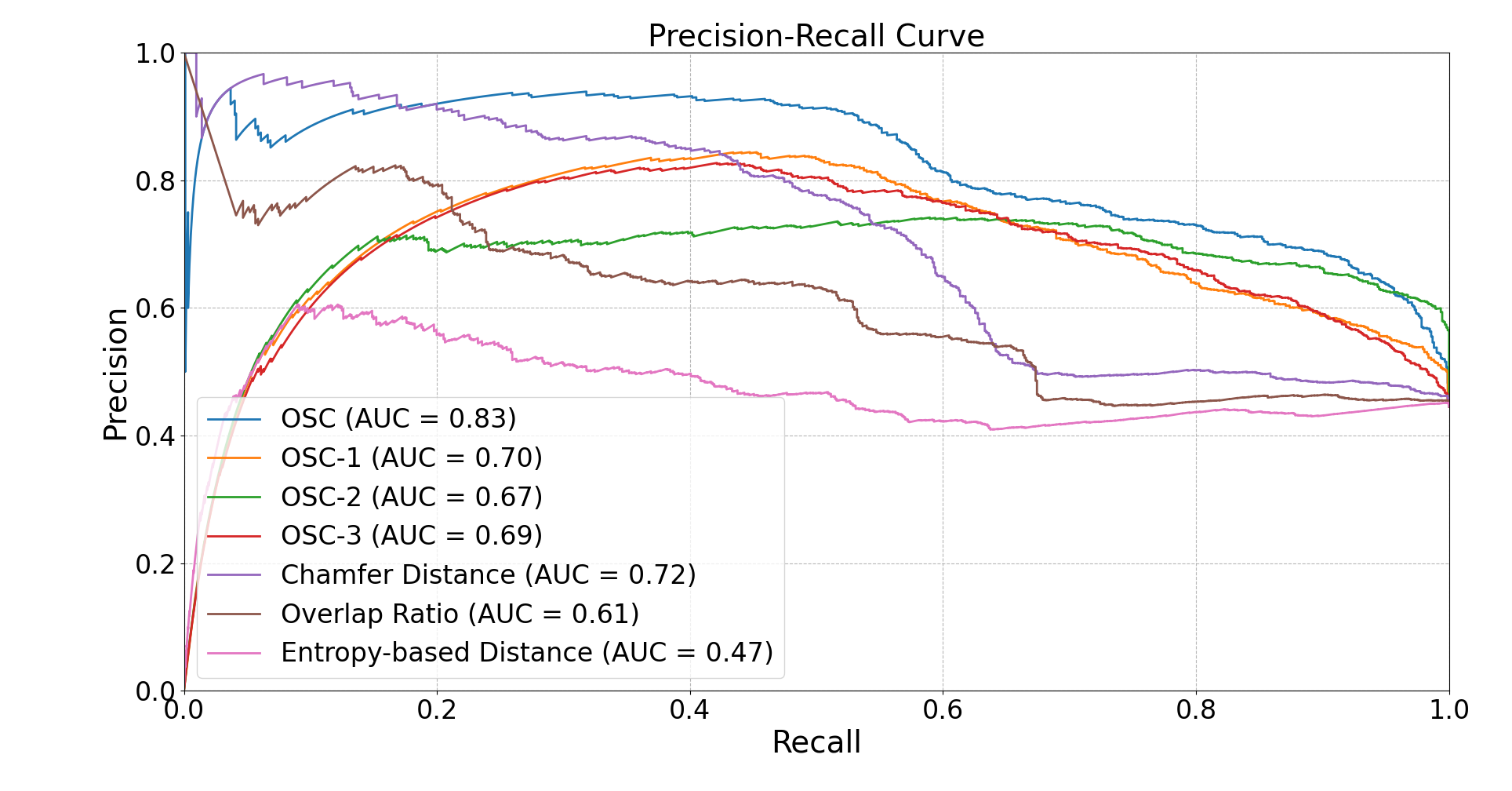}
\caption{Precision and recall curves of the proposed occupancy-aware confidence score compared with other variants and methods. OSC denotes the proposed confidence score. OSC-1 denotes the score considering only the award component. OSC-2 includes an additional award component where both the submap and BIM are free. OSC-3 introduces an additional penalty component where the submap is occupied but the BIM is free.}
\label{fig:pr}
\end{figure}

\subsection{Registration Reliability Assessment}
\label{exp:conf}

The proposed occupancy-aware confidence score is designed to assess the reliability of registration. To evaluate its effectiveness, we construct a new benchmark in addition to the LiBIM-UST dataset. This benchmark includes unregistrable pairs created by combining cross-floor submaps and BIM models. The confidence score is employed to determine whether the estimated transformation is correct. For instance, if the score is high for a correctly registered pair of submaps and BIMs, it is considered a true positive. Conversely, if the score is high for an unregistrable pair, it is classified as a false positive. To analyze its performance, we plot the precision-recall (PR) curves in Figure~\ref{fig:pr}.

The compared methods include the proposed occupancy-aware confidence score, the Chamfer distance (with distances truncated at 1 m), the overlap ratio~\cite{yin2019failure} (with a neighborhood radius of 1 m), and the entropy-based distance~\cite{adolfsson2021coral} (with a neighborhood radius of 1 m). Additionally, we consider two specific cases within the confidence score: (1) scenarios where both the submap and BIM are free, treated as an award component, and (2) scenarios where the submap is occupied but the BIM is free, treated as a penalty component. These cases, which were not included in the original score, are detailed in Section~\ref{sec:verify}. Moreover, we also evaluate the impact of removing the original penalty component. The results demonstrate that the proposed confidence score (OSC in Figure~\ref{fig:pr}) effectively distinguishes between correct and incorrect registrations, outperforming the compared methods.

\begin{figure*}[!ht]
  \centering
  \includegraphics[width=0.91\textwidth]{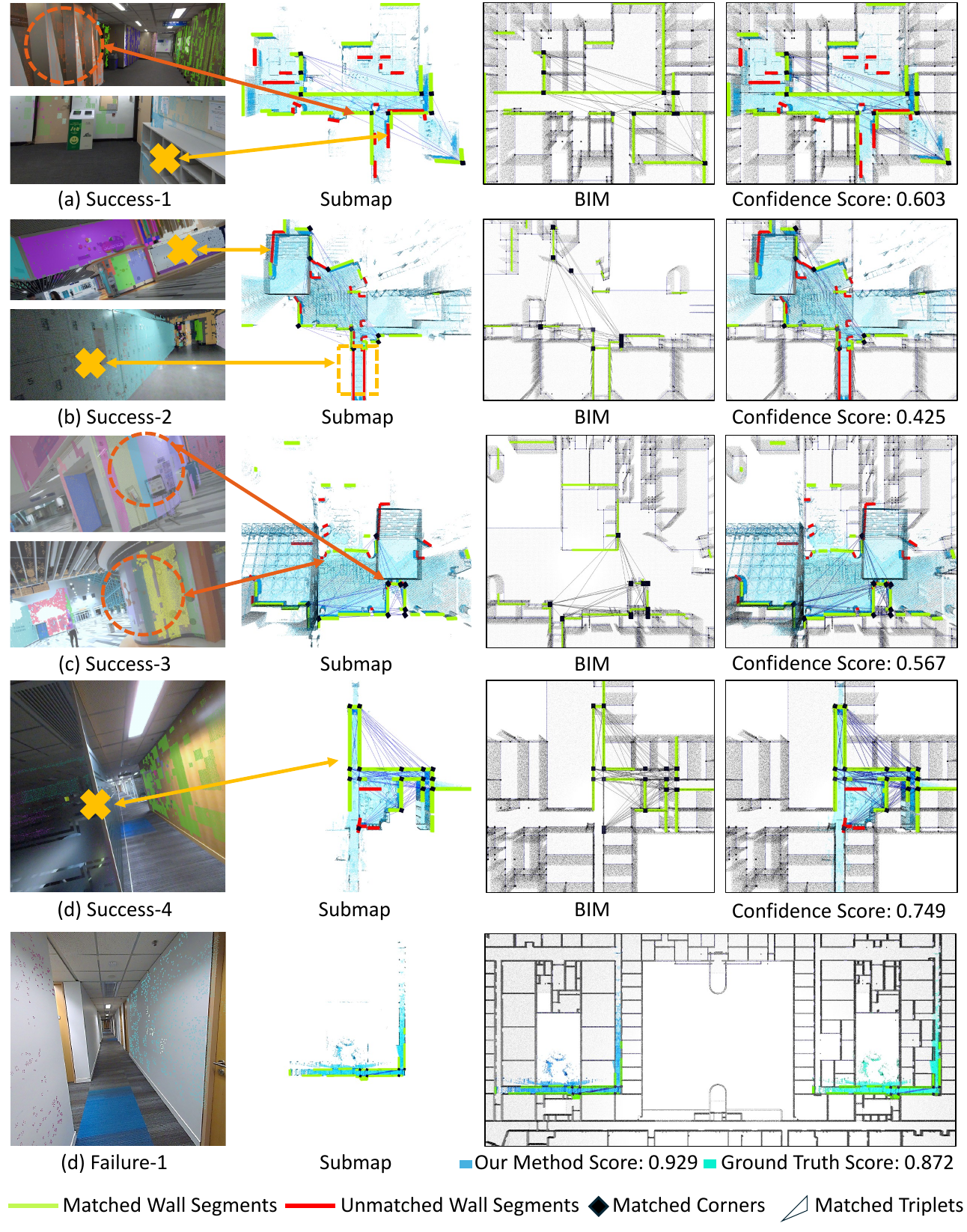}
  \caption{Featured scenarios of LiDAR submap-to-BIM registration are presented, including four success cases and one failure case. In each case, the following are shown: RGB images; the submap, highlighting matched walls in \textcolor{green}{green} and unmatched walls in \textcolor{red}{red}; the BIM with matched walls, corners, and triplets, and the registration result along with its confidence score. Two special types, construction inconsistencies and unstructured walls, are highlighted in yellow and brown, respectively. In \textbf{Success-1}, a jagged wall, marked by a brown circle, is identified as a plane and successfully matched. A yellow cross indicates an extra bookshelf, identified as an unmatched plane. In \textbf{Success-2}, the top yellow cross indicates a misaligned wall caused by a large slanted LED display that exists in the actual environment but is not present in the BIM model. The bottom yellow crosses indicate lockers on both sides of a corridor, which visually narrow the original corridor. In \textbf{Success-3}, brown circles highlight two curved walls that are segmented into several small planes and successfully matched. In \textbf{Success-4}, the yellow cross indicates a transparent glass partitioned in the real world, which corresponds to an unmatched solid wall in the BIM. The failure case, \textbf{Failure-1}, shows an ambiguous long corridor scenario, where even the ground truth transformation yields a lower confidence score.}
  \label{fig:casestudy}
\end{figure*}
% Case Study需要回复的意见
% 1. AE: LiDAR是非结构化的，BIM是结构化的，两个不匹配。建筑不完美是另一个问题。
% 2. 长走廊，曲面墙壁，或者特征比较少的区域。墙和角点可能不多。
% 3. Comment 3.4：墙体移除或者增加
% 4. Comment 4.2: 相似场景或者不同层

\subsection{Case Studies and Limitation Discussion}
\label{exp:case}
\subsubsection{Overview of Case Studies}
Figure \ref{fig:casestudy} illustrates several cases of LiDAR submap-to-BIM registration. To clarify the registration process, RGB images, extracted wall and corner features, constructed corner triplets, and the proposed confidence score are presented. Two special cases, construction inconsistencies and unstructured walls, are highlighted in yellow and brown, respectively.

\subsubsection{Construction Inconsistencies}
As shown in Figure~\ref{fig:casestudy}, construction inconsistencies typically manifest in three forms: unconstructed parts (i.e., elements present in the BIM model but missing in reality, such as the transparent glass in Success-4), additional constructed parts (i.e., elements absent in the BIM model but present in reality, such as the bookshelf in Success-1 and lockers in Success-2), and position biases (i.e., elements that exist in both but are misaligned, such as the top image in Success-2). While these inconsistencies increase the number of unmatched triplets, our method prioritizes finding a sufficient number of matched triplets rather than ensuring a perfect bijective correspondence between corners. Theoretically, we only need to construct at least one correct corner triplet correspondence for pose estimation, while the proposed confidence score ensures the identification of the correct transformation candidate during the verification stage. 

Furthermore, in computing the confidence score, scenarios where the submap is occupied but the BIM is free are not considered. This design ensures that while such inconsistencies, like additional constructed parts, do not contribute positively to the registration, they also do not negatively affect it in most cases. In summary, construction inconsistencies may impact the number of matched corner triplets and the quality of the confidence score. However, as long as the true transformation candidate outperforms others during the verification step, successful registration can still be achieved.

\subsubsection{Unstructured Walls}
Unstructured walls are represented as jagged walls (e.g., in Success-1) and curved walls (e.g., in Success-3) in the provided examples. Thanks to the plane threshold parameter $\sigma_{\lambda}$, our method does not require walls to be perfectly planar. Jagged walls are successfully detected as planes and matched accordingly. Curved walls are segmented into several small wall sections, which may produce several inaccurate corners. However, due to the design of our approach, which accounts for handling inaccurate corners through the discretization of descriptor and transformation spaces, curved walls still contribute positively to the registration process.

\subsubsection{Limitations of the Proposed Method}
While the above cases demonstrate the robustness of our method in handling challenging scenarios, there are still several limitations. First, our method requires at least one matched corner triplet and a sufficient number of matched walls to achieve a high confidence score. This means that larger submaps with more walls are more likely to result in successful registration. Second, although our method handles jagged and moderately curved walls effectively, it struggles with extreme cases, such as large curved walls, which fail to produce stable corners. Third, as illustrated in Failure-1, our method is affected by long corridor scenarios due to perceptual aliasing, a well-known systematic issue where two different locations exhibit similar appearances. To address this, additional information is needed to resolve ambiguities, such as enlarging the submap size. Alternatively, incorporating visual information could provide semantic cues, such as construction materials, functional zoning, or room divisions, which can be matched with BIM semantics to eliminate ambiguities.

\section{Conclusion and Future Work}
\label{sec:conclusion}
In this study, we propose a one-shot global registration method capable of aligning LiDAR data to BIM without requiring an initial guess. Walls and corners are extracted from both BIM and LiDAR data as shared features. Triangle descriptors are then computed based on corner triplets and matched using a Hash table, enabling fast and efficient retrieval independent of the BIM model's size. In the back end, we incorporate the Hough transform and a hierarchical voting mechanism to hypothesize multiple transformation candidates. Subsequently, we introduce an occupancy-aware score to identify the optimal pose estimation. In the LiBIM-UST dataset, we demonstrate the effectiveness of our method in comparison with traditional image and point cloud registration techniques.

% \revise{As shown in Figure~\ref{fig:casestudy}, our method still encounters challenges when deployed in complex indoor scenarios, such as long corridors and repetitive environments. We attribute this limitation to the inability of the proposed triangle descriptor to handle these scenarios effectively. To address these challenges, we propose two potential solutions: first, incorporating local descriptors to improve registration performance; and second, utilizing color information obtained from mapping results, as provided by our sensor suite shown in Figure~\ref{fig:sensor}, along with advanced sensor fusion techniques~\cite{lin2022r}.}

Future research will encompass both robotics and construction automation. In robotics, integrating global localization methods, such as the particle filter, presents a promising direction for enhancing applications that utilize BIM as a navigational map. We also plan to leverage the semantic information of BIM to enhance the performance. For construction automation, achieving global alignment between point clouds and BIM is not enough; it is also critical to identify dissimilar locations and elements to enable BIM correction for quality management. This capability is crucial for supporting AR applications on BIM.

% \yhcomment{comment 3.7 Future Research and Limitations:} \yhcomment{comment 4.4 also}

\section*{Aknowledgement}

We sincerely thank Prof. Jack Chin Pang Cheng for providing the BIM of the HKUST Academic Building.

% \yhcomment{leave it to yh}

\bibliographystyle{IEEEtran}
\bibliography{bare_jrnl}

% IEEE AUTHORS

\end{document}